\documentclass[letterpaper, 10 pt, conference]{ieeeconf}  % Comment this line out if you need a4paper

\IEEEoverridecommandlockouts                              % This command is only needed if 
\usepackage[algo2e]{algorithm2e} 
\usepackage{algorithm}
\usepackage{algpseudocode}
\algnewcommand\algorithmicinput{\textbf{Input:}}
\algnewcommand\INPUT{\item[\algorithmicinput]}
\usepackage{graphics} % for pdf, bitmapped graphics files
\usepackage{graphicx} % for pdf, bitmapped graphics files
\usepackage{times} % assumes new font selection scheme installed
\usepackage{amsmath} % assumes amsmath package installed
\usepackage{amssymb}  % assumes amsmath package installed
\usepackage{color}
\usepackage{subfig}
\usepackage{epstopdf}
\usepackage{wrapfig}
\usepackage{adjustbox}
\usepackage{booktabs}
\usepackage{multirow}
\usepackage{balance}
\usepackage{algpseudocode}
\usepackage{amsmath}

\usepackage{xcolor}
\usepackage{mdframed}
\usepackage{newfloat}
\DeclareFloatingEnvironment[fileext=frm,placement={!ht},name=Frame]{myfloat}
\makeatletter
\newenvironment{AlgoBox}[2]{ % Takes two arguments and not one
\begin{myfloat}[tb!]
\protected@edef\@currentlabelname{#1}
\protected@edef\@currentlabel{#2}
\begin{mdframed}[
innerlinewidth=0.5pt,
innerleftmargin=10pt,
innerrightmargin=10pt,
innertopmargin = 10pt,
innerbottommargin=10pt,
skipabove=\dimexpr\topsep+\ht\strutbox\relax,
roundcorner=5pt,
frametitle={#1},
frametitlerule=true,
frametitlerulewidth=1pt]
}{
\end{mdframed}
\end{myfloat}
}
\makeatother
\title{\LARGE \bf
Factored Pose Estimation of Articulated Objects using Efficient Nonparametric Belief Propagation  
}

\author{Karthik Desingh$^{1}$, Shiyang Lu$^{1}$, Anthony Opipari$^{1}$,  Odest Chadwicke Jenkins$^{1}$ % <-this % stops a space %Everyone should add their name when they are happy with the paper
\thanks{$^{1}$Department of Electrical Engineering and Computer Science, Robotics Institute, University of Michigan, Ann Arbor
        {\tt\small \{kdesingh,shiyoung,topipari,ocj\}@umich.edu}}
}
\begin{document}

\maketitle
\thispagestyle{empty}
\pagestyle{empty}

%%%%%%%%%%%%%%%%%%%%%%%%%%%%%%%%%%%%%%%%%%%%%%%%%%%%%%%%%%%%%%%%%%%%%%%%%%%%%%%%
\begin{abstract}
Robots working in human environments often encounter a wide range of articulated objects, such as tools, cabinets, and other jointed objects. 
%These objects can articulate their parts based on joint constraints to produce different states/poses. 
Such articulated objects can take an infinite number of possible poses, as a point in a potentially high-dimensional continuous space.  A robot must perceive this continuous pose to manipulate the object to a desired pose. This problem of perception and manipulation of articulated objects remains a challenge due to its high dimensionality and multi-modal uncertainty.  
%In particular, problem which grows with the number of objects and is multi-modal in nature with partial and ambiguous observations of the real world. 
In this paper, we propose a factored approach to estimate the poses of articulated objects using an efficient nonparametric belief propagation algorithm. We consider inputs as geometrical models with articulation constraints, and observed RGBD sensor data. The proposed framework produces object-part pose beliefs iteratively. The problem is formulated as a pairwise Markov Random Field (MRF) where each hidden node (continuous pose variable) is an observed object-part's pose and the edges denote the articulation constraints between the parts. We propose articulated pose estimation by Pull Message Passing algorithm for Nonparametric Belief Propagation (PMPNBP) and evaluate its convergence properties over scenes with articulated objects. 
\end{abstract}
\IEEEpeerreviewmaketitle

%%%%%%%%%%%%%%%%%%%%%%%%%%%%%%%%%%%%%%%%%%%%%%%%%%%%%%%%%%%%%%%%%%%%%%%%%%%%%%%%
\begin{figure*}[ht!]
    \centering
\subfloat[Robot observing a cabinet with drawers]{\includegraphics[width=0.28\textwidth]{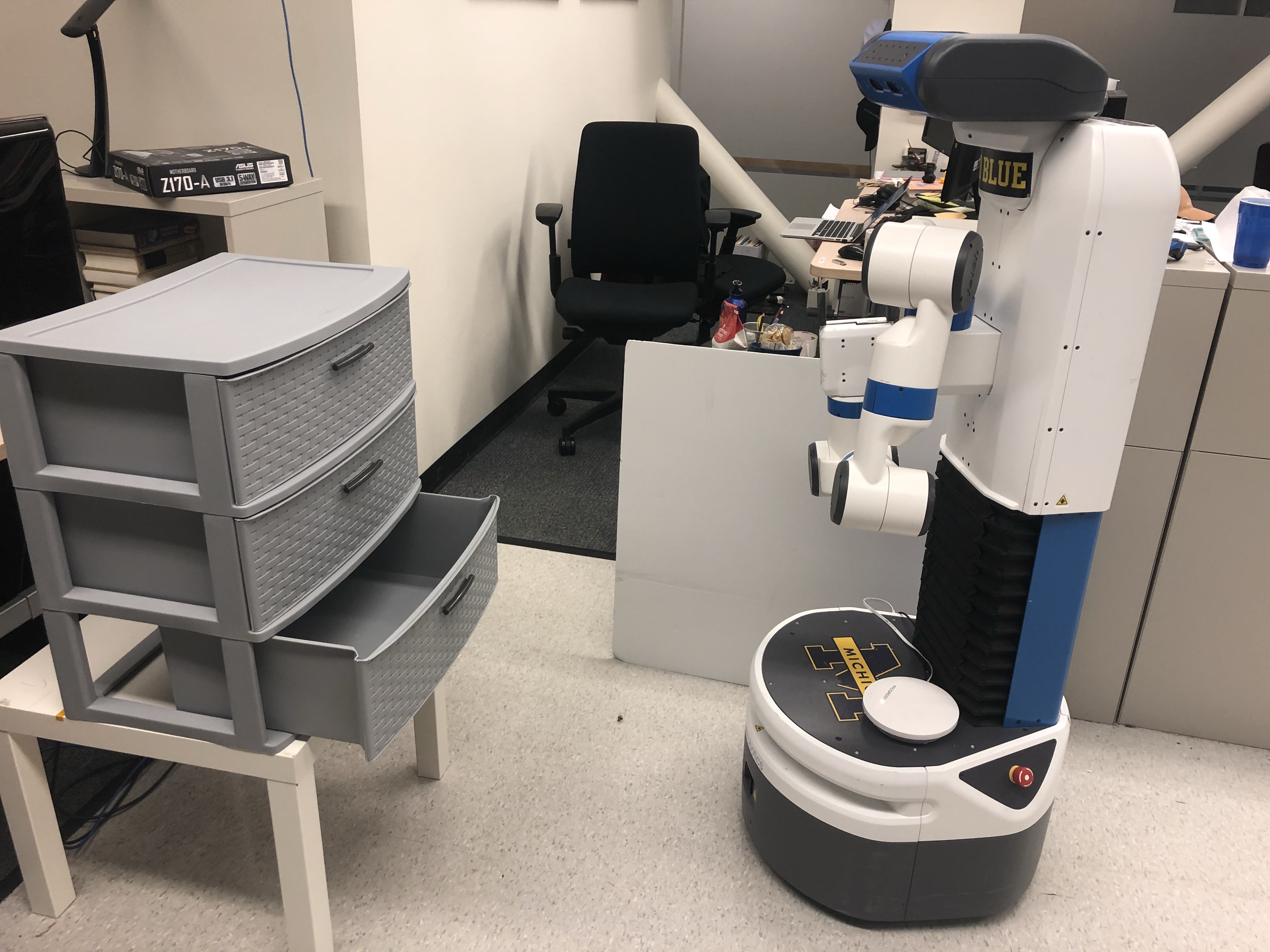}}~
\hspace{0.3cm}
%\begin{adjustbox}{minipage=0.2\textwidth}
\subfloat[Point cloud observation]{\includegraphics[width=0.28\textwidth]{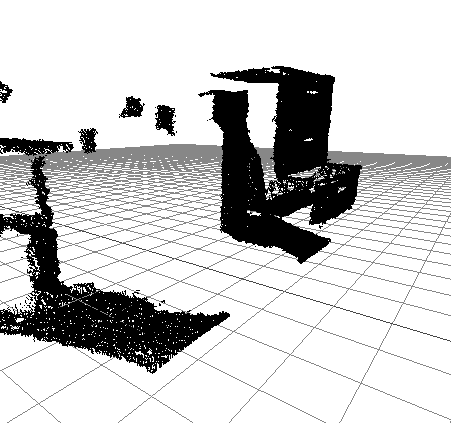}}~ \\
%\hspace{-0.4cm}

\medskip
\subfloat[Belief at iteration 0]{\includegraphics[width=0.28\textwidth]{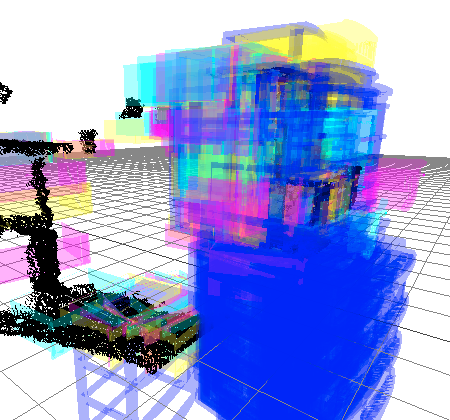}}~
\hspace{0.3cm}
\subfloat[Belief at iteration 100]{\includegraphics[width=0.28\textwidth]{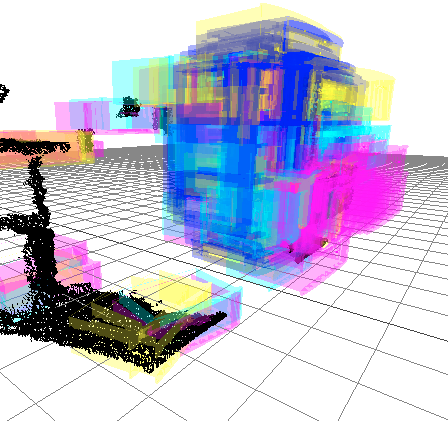}}~
\hspace{0.3cm}
\subfloat[Maximum likely estimate]{\includegraphics[width=0.28\textwidth]{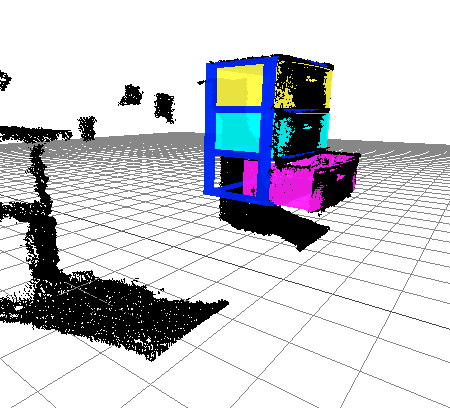}}\\
% \end{adjustbox}
% \subfloat[Belief on tools at iteration x]{\includegraphics[width=0.475\columnwidth]{./figures/blank.jpg}}~
% \subfloat[Belief on cabinet at iteration x]{\includegraphics[width=0.475\columnwidth]{./figures/blank.jpg}}\\
% \subfloat[Estimated pose of tools ]{\includegraphics[width=0.475\columnwidth]{./figures/blank.jpg}}~
% \subfloat[Estimated pose of cabinet ]{\includegraphics[width=0.475\columnwidth]{./figures/blank.jpg}}
 
 \caption{\footnotesize Robot estimating the state of a cabinet with 3 prismatically articulated drawers from a 3D point cloud.}
 \label{teaser}
\end{figure*}

\section{Introduction}

%\kar{Perception in robotics- its importance - connecting to task level planning}\kar{What is the problem being addressed}

%Robots perform tasks - that has sequence of manipulation across various objects - articulated objects - this paper focuses on the perception of these objects. These objects can be categorized as furnitures (chairs, tables), storage objects (tool chest, cabinets, fridge) and small objects (plier, coke can, detergent bottle). Similarly a fridge can be composed of two rigid bodies (door and container) that articulate to open and close the storage location. 
%Personal robots must be able to perform tasks reliably using various household objects that prevail in human environment. 

Robots working in human environments often encounter a wide range of articulated objects, such as tools, cabinets, and other kinematically jointed objects. 
%Many of these objects are designed to articulate kinematically.  
For example, the cabinet with three drawers shown in Fig~\ref{teaser} functions as a storage container. An robot would need to perform open and close actions on the drawers to accomplish storage and retrieval tasks.  Accomplishing such tasks involves repeated sense-plan-act phases under uncertainty in the robot's observations and demands pose estimation that accommodates uncertainty to inform a planner with current state of the world. This uncertainty poses the challenge of dealing with sensor noise and inherent environmental occlusions. 
%More specifically, to be able to perform a storage or a retrieval task using this cabinet, robot should be able to perceive not only the cabinet's continuous pose in the world, but its drawers and their handles. 
Ability to perceive articulated pose under partial observations due to self and environmental occlusions makes the inference problem multi-modal. Further, the inference becomes a high-dimensional problem when the number of object parts grow in number. 

Pose estimation methods have been proposed that take a generative approach to the problem~\cite{sui2017goal,desingh2016physically,zeng2018semantic}.  These methods aim to explain a scene as a collection of object/parts poses, using a particle filter formulation to iteratively maintain belief over possible states in the form of particles. Though these approaches hold the power of modeling the world generatively, they have a inherent drawback of being slow with the increase in the number of rigid bodies. In this paper, we focus on overcoming this drawback by factoring the state as individual object parts constrained by their articulations to create an efficient inference framework for pose estimation.
%Interactively perceiving, learning and understanding the articulations on a novel object using either human demonstrations or robot manipulations \cite{martin14online,martin2016intg,hausman2015artic,sturm11prob,sturm13book} is a largely explored direction. In this paper we propose a method that exploits such articulation models to aid an efficient inference mechanism. 

%Generative approaches - DPM - articulated human parts - NBP - Our proposed method
%Generative approaches have been proposed in the past to estimate object poses on cluttered scenes [CITE all relevant work] with assumption that objects are rigid bodies. Other works such as [CITE all relevant work] divide objects into their parts to perform recognition and pose estimation, with assumption of no articulations. 
Generative methods exploiting articulation constraints are widely used in human pose estimation problems~\cite{sigal2004tracking, sudderth2004visual,vondrak2013dynamical} where human body parts have constrained articulation. We take a similar approach and factor the problem using a Markov Random Field (MRF) formulation where each hidden node in the probabilistic graphical model represents an observed object-part's pose (continuous variable), each observed node has the information about the object-part from observation and edges of the graph denote the articulation constraints between the parts. Inference on the graph is performed using a message passing algorithm that share the information between the parts' pose variables, to produce their pose beliefs which collectively gives the state of the articulated object.
 
Existing message passing approaches~\cite{isard2003pampas,sudderth2003nonparametric} represent message as a mixture of Gaussian components and provide Gibbs sampling based techniques to approximate message product and update operations. Their message representation and product techniques limits the number of samples used in the inference and is not applicable to our application domain. In this paper we provide a more efficient ``pull'' Message Passing algorithm for Nonparametric Belief Propagation (PMPNBP).  The key idea of pull message updating is to evaluate samples taken from the belief of the receiving node with respect to the densities informing the sending node.  The mixture product approximation can then be performed individually per sample, and then later normalized into a distribution.  This pull updating avoids the computational pitfalls of push updating of message distributions used in~\cite{isard2003pampas,sudderth2003nonparametric}.

Our system  takes in 3D point cloud as the sensor data and object geometry models in the form of an URDF (Unified Robot Description Format) as input and outputs belief samples in continuous pose domain. We use these belief samples to compute a maximum likely estimate to let the robot act on the object. We evaluate the performance of the system by quantifying over an articulated object on compelling scenes. Contributions of this paper include: a) proposal of an efficient belief propagation algorithm to estimate articulated object poses, b) discussion and comparisons with traditional particle filter as baseline, c) a belief representation from perception to inform a task planner. A simple task is illustrated to show how the belief propagation informs a task planner to choose an information gain action and overcome uncertainty in the perceptual estimation.

\section{Related work}
%(TODO - include Dual-Quaternion works and interpolation in Quaternions

Existing methods in the literature have set out to address the challenge of manipulating articulated objects by robots in complex human environments. Particular focus has been placed on addressing the task of estimating novel articulated objects' kinematic models by a robot through interactive perception. Hausman et al.~\cite{hausman2015artic} propose a particle filtering approach to estimate articulation models and plan actions to reduce the model uncertainty. In~\cite{martin14online}, Martin et al. suggest an online interactive perception technique for estimating kinematic models by incorporating low-level point tracking and mid-level rigid body tracking with high-level kinematic model estimation over time. Sturm et al.~\cite{sturm11prob,sturm13book} addressed the task of estimating articulation models in a probabilistic fashion by human demonstration of manipulation examples. 
%Katz et al.~\cite{katz10intr} also pair a RANSAC-based plane fitting algorithm and iterative rectangle search algorithm to perform marker-less pose estimation; however, they do not leverage their previously learned kinematic models. .

All of these approaches discover the articulated object's kinematic model by alternating between action and sensing and are important methods to enable a robot is to reliably interact with novel articulated objects. In this paper we assume that such kinematic models once learned for an object can be reused to localize their articulated pose under real world ambiguous observations. The method proposed in this paper could compliment the existing body of work towards task completion in unstructured human environment.

Probabilistic graphical model representations such as Markov random field (MRF) are widely used in computer vision problems where the variables take discrete labels such as foreground/background. Many algorithms have been proposed to compute the joint probability of the graphical model. Belief propagation algorithms are guaranteed to converge on tree-structured graphs. For graph structures with loops, Loopy Belief Propagation (LBP)~\cite{murphy1999loopy} is empirically proven to perform well for discrete variables. The problem becomes non-trivial when the variables take continuous values. Sudderth et.al (NBP)~\cite{sudderth2003nonparametric} and Particle Message Passing (PAMPAS) by Isard et.al~\cite{isard2003pampas} provide sampling approaches to perform belief propagation with continuous variables. Both of these approaches approximate a continuous function as a mixture of weighted Gaussians and use local Gibbs sampling to approximate the product of mixtures. NBP has been effectively used in applications such as human pose estimation~\cite{sigal2004tracking} and hand tracking~\cite{sudderth2004visual} by modelling the graph as a tree structured particle network. Scene understanding problems where a scene is composed of household objects with articulations demands large number of samples in the representation to handle the high-dimensional multimodal state space. The algorithm proposed in this paper produces promising results to handle such demands. We reported comparisons with existing NBP algorithm ~\cite{isard2003pampas} in \cite{desingh18pmpnbp} with 2D examples.

Model based generative methods~\cite{narayanan2016discriminatively,sui2017sum,xiang2017posecnn} are increasingly being used to solve scene estimation problems where heuristics from discriminative approaches~\cite{NIPS2015_5638, girshick2014rich} are used to infer object poses. These approaches do not account for object-object interactions or articulations and rely significantly on the effectiveness of recognition. Our framework doesn't rely on any prior detections but can benefit from them while inherently handling noisy priors~\cite{sudderth2003nonparametric,isard2003pampas,desingh18pmpnbp}. Chua et. al~\cite{chua2016scene} proposed a scene grammar representation and belief propagation over factor graphs, whose objective similar to ours for generating scenes with multiple-objects satisfying the scene grammars. This approach is similar to ours however, we specifically deal with 3D observations along with continuous variables.

%Recent works such as~\cite{xu2017scene} have proposed systems that can work on wild image data and refine the object detections along with their relations. However, these methods do not consider the continuous pose in their estimation and work in pixel domain. 

%\input{motivation.tex}
\section{Problem Statement} \label{problem_statement}
We consider an articulated object $O$ to be comprised of with $N$ object-parts and $N-1$ points of articulation.  Such an object description conforms with the Unified Robot Description Format (URDF) commonly used in the Robot Operating System (ROS)~\cite{ros}.  Such a URDF-compliant kinematic model can be represented using an undirected graph $G=(V, E)$ with nodes $V$ for object-part links and edges $E$ for points of articulation. If $G$ is a Markov Random Field (MRF), it has two types of variables $X$ and $Y$ that are, respectively, hidden and observed variables. Let $Y=\{Y_s \mid Y_s \in V\}$, where $Y_s = P_s \subseteq P$, with $P$ being the point cloud observed by the robot's 3D sensor. Each object-part has an observed node in the graph $G$. $P_s$ serves as a region of interest if a trained object detector is used to find the object in the scene, but is optional in our current approach. Each observed node $Y_s$ is connected to a hidden node $X_s$ that represents the pose of the underlying object part. Let $X=\{X_s\mid X_s \in V\}$, where $X_s \in \mathbb{H}_D$ is a dual quaternion pose of an object-part. Dual quaternions~\cite{gilitschenski2014new,kenwright2012beginners} are a quaternion equivalent to dual numbers representing a 6D pose $X_s=(x, y, z, q_w, q_x, q_y, q_z)$ as $X_s = q_r + \epsilon q_d$ where $q_r$ is the real component and $q_d$ is the dual component. Alternatively it is represented as $X_s = [q_r][q_d]$. Constructing a dual quaternion $X_s$ is similar to rotation matrices, with a product of dual quaternions representing translation and orientation as $X_s=dq_{pos}*dq_{ori}$, where $*$ is a dual quaternion multiplication. $dq_{ori} = [q_w, q_x, q_y, q_z][0,0,0,0]$ is the dual quaternion representation of pure rotation and $dq_{pos} = [1, 0, 0, 0][0,\frac{x}{2},\frac{y}{2},\frac{z}{2}]$ is the dual quaternion representation of pure translation. This dual quaternion representation is widely used for rigid body kinematics, where the $*$ operation due to its efficiency and elegance compared with matrix multiplication. In addition to the representing the hidden variable $X_s$, dual quaternions can capture the constraints in the edges $E$ and represent articulation types such as prismatic, revolute, and fixed effectively. This will be discussed in detail in Section \ref{pairwise}.

\begin{figure}
    \centering
    \captionsetup[subfigure]{labelformat=empty}
	\subfloat{\includegraphics[width=0.8\columnwidth]{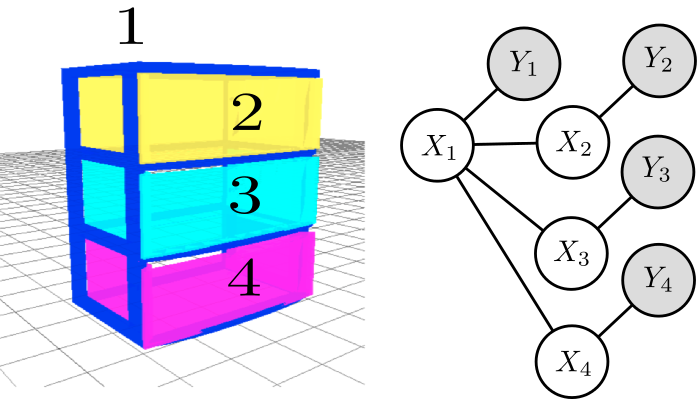}}
 \caption{\footnotesize Cabinet with 3 drawers connect to its frame is converted to a probabilistic graphical model with hidden nodes $X_s$ representing the pose of the object-parts and observed nodes $Y_s$ connected to each of the hidden nodes.}
 \label{problem_statement}
\end{figure}

Pose estimation of the articulated object involves inferring the hidden variables $X_s$ that maximizes the joint probability of the graph $G$ considering only second order cliques, which is given as: 
\begin{equation}\label{eq:1}
p( X, Y ) = \frac{1}{Z}\prod_{(s, t)\in E} \psi_{s, t}(X_s, X_t) \prod_{s \in V} \phi_s(X_s, Y_s)
\end{equation}
where $\psi_{s, t}(X_s, X_t)$ is the pairwise potential between nodes $X_s$ and $X_t$, $\phi_s(X_s, Y_s)$ is the unary potential between the hidden node $X_s$ and observed node $Y_s$, and $Z$ is a normalizing factor. The problem is to infer belief over the possible articulation poses assigned to hidden variables $X$ that are continuous such that the joint probability is maximized. This inference is generally performed by passing messages between hidden variables $X$ until convergence of their belief distributions over several iterations. After converging over iterations, a maximum likelihood estimate of the marginal belief gives the pose estimate $X_s^{est}$ of a object-part corresponding to the node in the graph $G$. The collection of all such object-part pose estimates form the entire object's pose estimate.

\section{Nonparametric Belief Propagation}
% \begin{figure*}
%     \centering
%     \captionsetup[subfigure]{labelformat=empty}
% 	\subfloat{\includegraphics[width=0.7\textwidth]{./figures/pmpnbp_flow_chart.eps}}
%  \caption{\footnotesize Flow of the ``pull'' message passing: This figure aids the algorithms presented in this section.}
%  \label{msg_pull_push}
% \end{figure*}

\subsection{Overview}
\label{nbp_overview}
A message is denoted as $m_{t\rightarrow s}$ directed from node $t$ to node $s$ if there is an edge between the nodes in the graph $G$. The message represents the distribution of what node $t$ thinks node $s$ should take in terms of the hidden variable $X_s$. Typically, if $X_s$ is in the continuous domain, then $m_{t\rightarrow s}(X_s)$ is represented as a Gaussian mixture to approximate the real distribution: 
\begin{equation}\label{eq:2}
m_{t\rightarrow s}(X_s) = \sum_{i=1}^M w_{ts}^{(i)} \mathcal{N}(X_s; \mu_{ts}^{(i)}, \Lambda_{ts}^{(i)})
\end{equation}
where $\sum_{i=1}^M w_{ts}^{(i)} = 1$, $M$ is the number of Gaussian components, $w_{ts}^{(i)}$ is the weight associated with the $i^{th}$ component, $\mu_{ts}^{(i)}$ and $\Lambda_{ts}^{(i)}$ are the mean and covariance of the $i^{th}$ component, respectively. We use the terms components, particles and samples interchangeably in this paper. Hence, a message can be expressed as $M$ triplets:

\begin{equation}\label{eq:3}
m_{t\rightarrow s} = \{(w_{ts}^{(i)}, \mu_{ts}^{(i)}, \Lambda_{ts}^{(i)}): 1\leq i \leq M\}
\end{equation}

\begin{AlgoBox}{\small Algorithm - Message update}{Algorithm - Message update}\label{alg:message_update}
\small
Given input messages $m_{u\rightarrow t}^{n-1}(X_t) = \{(\mu_{ut}^{(i)}, w_{ut}^{(i)})\}_{i=1}^{M}$ for each $u \in \rho(t)\setminus s$, and methods to compute functions $\psi_{ts}(X_t, X_s)$ and $\phi_t(X_t, Y_t)$ point-wise, the algorithm computes $m_{t\rightarrow s}^{n}(X_s)=\{(\mu_{ts}^{(i)}, w_{ts}^{(i)})\}_{i=1}^M$
\begin{enumerate}
    \item[1.] Draw $M$ independent samples $\{\mu_{ts}^{(i)}\}_{i=1}^M$ from $bel_s^{n-1}(X_s)$.
    \begin{enumerate}
        \item [(a)] If $n=1$ the $bel_s^{0}(X_s)$ is a uniform distribution or informed by a prior distribution.
        \item [(b)] If $n>1$ the $bel_s^{n-1}(X_s)$ is a belief computed at $(n-1)^{th}$ iteration using importance sampling.  %This gives us $M$ samples that forms which are not weighted/equally weighted.
    \end{enumerate}
    \item[2] For each $\{\mu_{ts}^{(i)}\}_{i=1}^M$, compute $w_{ts}^{(i)}$
    \begin{enumerate}
    \item[a] Sample $\hat{X}_t^{(i)} \sim \psi_{ts}(X_t, X_s=\mu_{ts}^{(i)})$ 
    \item[b] Unary weight $w_{unary}^{(i)}$ is computed using $\phi_t(X_t=\hat{X}_t^{(i)}, Y_t)$.
    \item[c] Neighboring weight $w_{neigh}^{(i)}$ is computed using $m_{u\rightarrow t}^{n-1}$.
    \begin{enumerate}
        \item[(i)] For each $u \in \rho(t) \setminus s$ compute $W_u^{(i)}=\sum_{j=1}^Mw_{ut}^{(j)}w_{u}^{(ij)}$ where \\ $w_{u}^{(ij)} = \psi_{ts}(X_s=\mu_{ts}^{(i)}, X_t=\mu_{ut}^{(j)})$.
        
        \item[(ii)] Each neighboring weight is computed by $w_{neigh}^{(i)}=\prod_{u \in \rho(t) \setminus s}W_{u}^{(i)}$
    \end{enumerate}
    \item[d] The final weights are computed as $w_{ts}^{(i)}=w_{neigh}^{(i)} \times w_{unary}^{(i)}$. 
    \end{enumerate}
    \item[3] The weights $\{w_{ts}^{(i)}\}_{i=1}^M$ are associated with the samples $\{\mu_{ts}^{(i)}\}_{i=1}^M$ to represent $m_{t\rightarrow s}^n(X_s)$.
\end{enumerate}
\end{AlgoBox}

Assuming the graph has tree or loopy structure, computing these message updates is nontrivial computationally.
A message update in a continuous domain at an iteration $n$ from a node $t \rightarrow s$ is given by
%\begin{equation}
\begin{multline}\label{eq:4}
m_{t\rightarrow s}^n(X_s) \leftarrow \\ \int_{X_t \in \mathbb{H}_D} \bigg(\psi_{st}(X_s, X_t)\phi_t(X_t, Y_t) \prod_{u \in \rho(t)\setminus s} m_{u\rightarrow t}^{n-1}(X_t)\bigg)dX_t 
\end{multline}
%\end{equation}
where $\rho(t)$ is a set of neighbor nodes of $t$. The marginal belief over each hidden node at iteration $n$ is given by
\begin{equation}\label{eq:5}
\begin{aligned}
bel_s^n(X_s) \propto \phi_s(X_s, Y_s) \prod_{t \in \rho(s)} m_{t\rightarrow s}^n(X_s) \\
bel_s^n = \{(w_s^{(i)}, \mu_s^{(i)}, \Lambda_s^{(i)}): 1\leq i \leq T\}
\end{aligned}
\end{equation}
where $T$ is the number of components used to represent the belief.

\subsection{``Push'' Message Update}
NBP~\cite{sudderth2003nonparametric} provides a Gibbs sampling approach to compute an approximation of the product $\prod_{u \in \rho(t)\setminus s} m_{u\rightarrow t}^{n-1}(X_t)$. 
Assuming that $\phi_t(X_t, Y_t)$ is pointwise computable, a ``pre-message'' \cite{ihler2009particle} is defined as
\begin{equation}\label{eq:6}
M_{t\rightarrow s}^{n-1}(X_t)=\phi_t(X_t, Y_t)\prod_{u \in \rho(t)\setminus s} m_{u\rightarrow t}^{n-1}(X_t)
\end{equation}
which can be computed in the Gibbs sampling procedure. This reduces Equation~\ref{eq:4} to
\begin{equation}\label{eq:7}
%\begin{multline}
m_{t\rightarrow s}^n(X_s) \leftarrow \int_{X_t \in \mathbb{R}^b}  \bigg(\psi_{st}(X_s, X_t)M_{t\rightarrow s}^{n-1}(X_t)\bigg)dX_t
%\end{multline}
\end{equation}

\begin{AlgoBox}{\small Algorithm - Belief update}{Algorithm - Belief update}\label{alg:belief_update}
\small
Given incoming messages $m_{t\rightarrow s}^n(X_t) = \{( w_{ts}^{(i)},\mu_{ts}^{(i)})\}_{i=1}^{M}$ for each $t \in \rho(s)$, and methods to compute functions $\phi_s(x_s, y_s)$ point-wise, the algorithm computes $bel_s^n(X_s) \propto \phi_s(X_s, Y_s) \prod_{t \in \rho(s)} m_{t\rightarrow s}^n(X_s) = \{( w_{s}^{(i)},\mu_{s}^{(i)})\}_{i=1}^{T}$
\begin{enumerate}
	\item[1] For each $t \in \rho(s)$
    \begin{enumerate}
    	\item[a] Update weights $w_{ts}^{(i)} = w_{ts}^{(i)} \times \phi(X_s=\mu_{ts}^{(i)}, Y_s)$.
        \item[b] Normalize the weights such that $\sum_{i=1}^M w_{ts}^{(i)} = 1$.
    \end{enumerate}
    \item[2] Combine all the incoming messages to form a single set of samples and their weights $\{(w_{s}^{(i)}, \mu_s^{(i)})\}_{i=1}^T$, where $T$ is the sum of all the incoming number of samples.
    \item[3] Normalize the weights such that $\sum_{i=1}^Tw_{s}^{(i)}=1$.
    \item[4] Perform a resampling step followed by diffusion with Gaussian noise, to sample new set $\{\mu_s^{(i)}\}_{i=1}^T$ that represent the marginal belief of $X_s$.
\end{enumerate}
\end{AlgoBox}

%The pairwise term $\psi_{st}(X_s, X_t)$ can be approximated as the marginal influence function $\zeta(X_t)$ to make the right side of Equation~\ref{eq:7} independent of $X_s$. %The marginal influence function provides the influence of $X_s$ for sampling $X_t$. 
%However, this function can be ignored if the pairwise potential function is based on the distance between the variables. This assumption makes Equation~\ref{eq:7} avoid the step of integration and 
NBP\cite{sudderth2003nonparametric} sample $\hat{X}_t^{(i)}$ from the ``pre-message'' followed by a pairwise sampling where $\psi_{st}(X_s, X_t)$ is acting as $\psi_{st}(X_s | X_t=\hat{X}_t^{(i)})$ to get a sample $\hat{X}_s^{(i)}$. 
%To represent message $m_{t\rightarrow s}^n(X_s)$, the $M$ samples $\{\hat{X}_s\}_{i=1}^M$ are considered as $\{\mu_{ts}\}_{i=1}^M$. $\{\Lambda_{ts}\}_{i=1}^M$ are computed using Kernel Density Estimation methods. PAMPAS~\cite{isard2003pampas} has a slightly different notation and methods to compute the samples. 

% \begin{mdframed}[frametitle={\small Algorithm - Overall Belief Propagation}]
% \label{alg:pmpnbp_overall}
%     \small
%     Given node potentials $\phi(X_s, Y_s)$ $\forall s \in V$, pairwise potentials $\psi(X_s, X_t)$ $\forall (s, t) \in E$ and initial messages for every edge $m_{s\rightarrow t}^0$ $\forall (s,t) \in E$, the algorithm iteratively updates all messages and computes the belief till the graph $G$ till converges.
%     \begin{enumerate}
%     	\item [1]  For $n \in [1:$max iterations$]$        
%         \begin{enumerate}
%         	\item [(a)] \textbf{Message update:}
%     \newline Update messages from iteration $(n-1)$ to $n$ using \textbf{Algorithm - Message Update}.
%     		\item [(b)] \textbf{Belief update:}
%      \newline Compute belief at iteration $n$ using messages at $n$ as described in \textbf{Algorithm - Belief Update}.
%         \end{enumerate}
%     \end{enumerate}
% \vspace{1mm}
% \end{mdframed}

%\chad{make sure algoboxes are aligned to the top of the text column.  you can use a minipage environment to ensure this layout.}

The Gibbs sampling procedure in itself is an iterative procedure and hence makes the computation of the "pre-message" (as the Foundation function described for PAMPAS) expensive as $M$ increases. 
%In the next section, we provide our proposed message representation followed by the algorithm to compute $m_{t\rightarrow s}^n(X_s)$ at iteration $n$.

% The marginal influence function is given by:
% \begin{equation}\label{eq:7}
% \zeta(X_t) = \int_{X_s \in \mathbb{R}^d} \psi_{st}(X_s, X_t) dX_s
% \end{equation}

% If the marginal influence function is also point wise computable then the entire product $\zeta(X_t)M_{t\rightarrow s}^{n-1}(X_t)$ can be computed as part of the sampling procedure proposed in NBP. Refer to the papers describing  NBP~\cite{sudderth2003nonparametric} and PAMPAS~\cite{isard2003pampas} for further details on how changes to the nature of the potentials affect the message update computation (as in Eq~\ref{eq:3}). 

\begin{figure*}[t!]
    \centering
    \captionsetup[subfigure]{labelformat=empty}
	\subfloat{\includegraphics[width=0.19\textwidth]{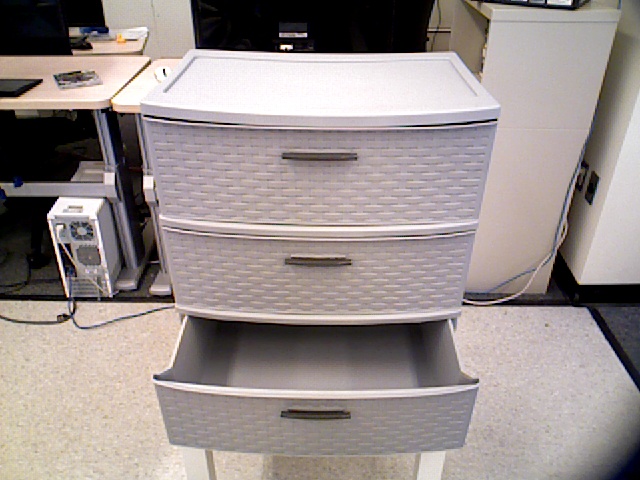}} ~
    \subfloat{\includegraphics[width=0.19\textwidth]{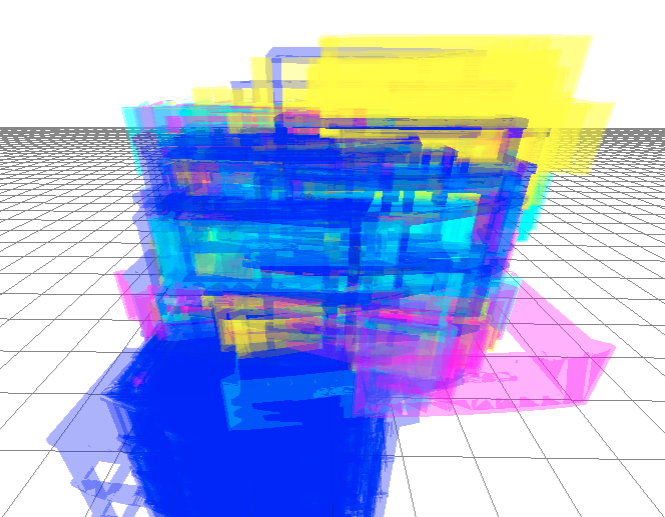}} ~
    \subfloat{\includegraphics[width=0.19\textwidth]{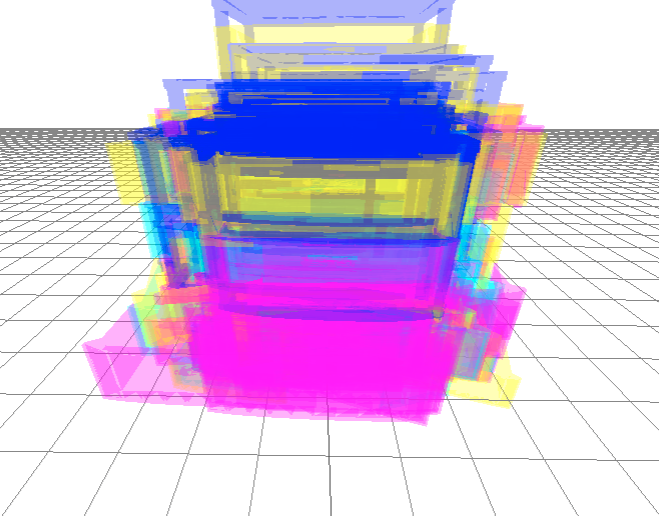}} ~
    \subfloat{\includegraphics[width=0.19\textwidth]{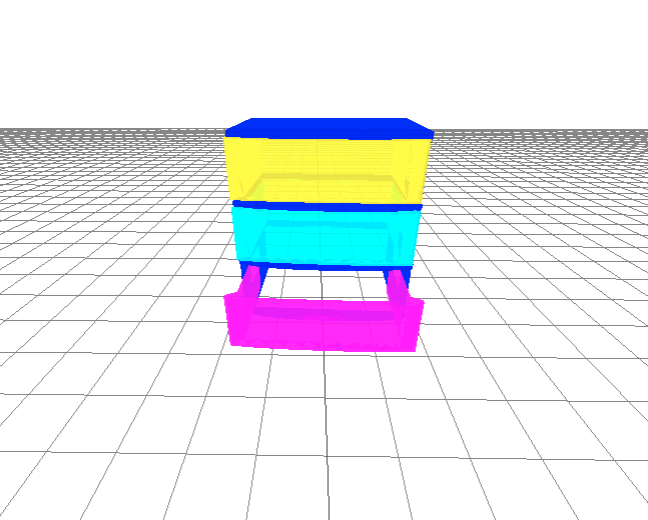}} ~
    \subfloat{\includegraphics[width=0.19\textwidth]{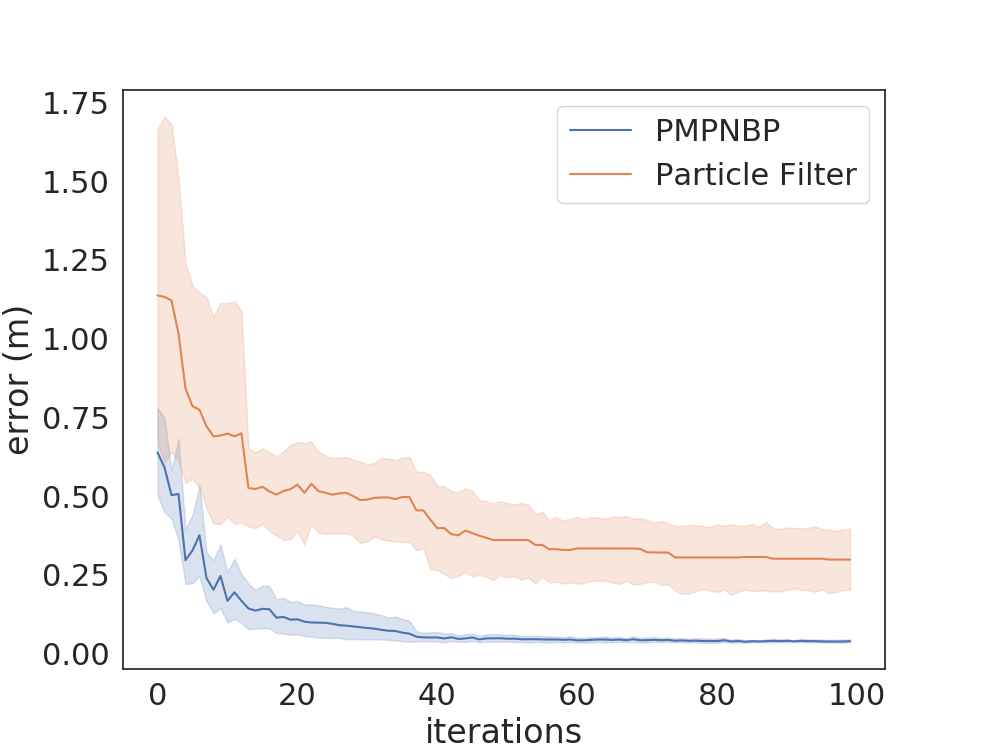}} \\
    
	\subfloat{\includegraphics[width=0.19\textwidth]{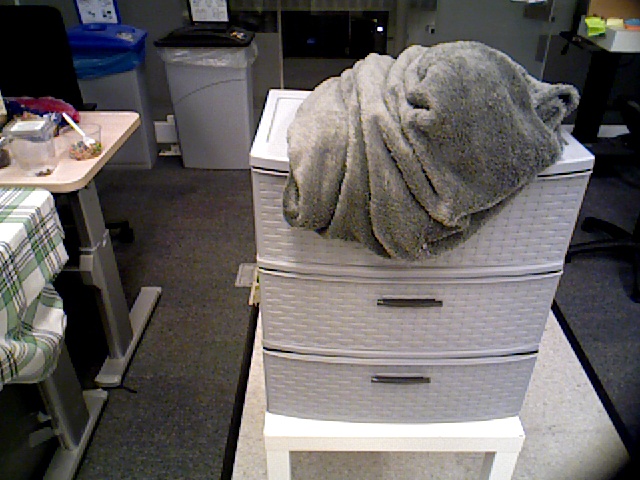}} ~
    \subfloat{\includegraphics[width=0.19\textwidth]{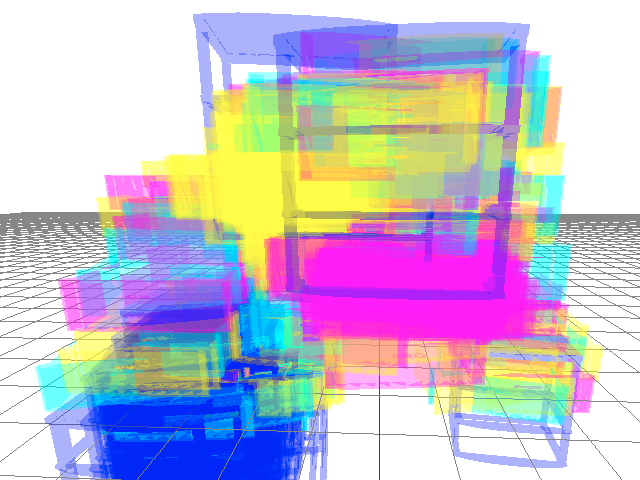}} ~
    \subfloat{\includegraphics[width=0.19\textwidth]{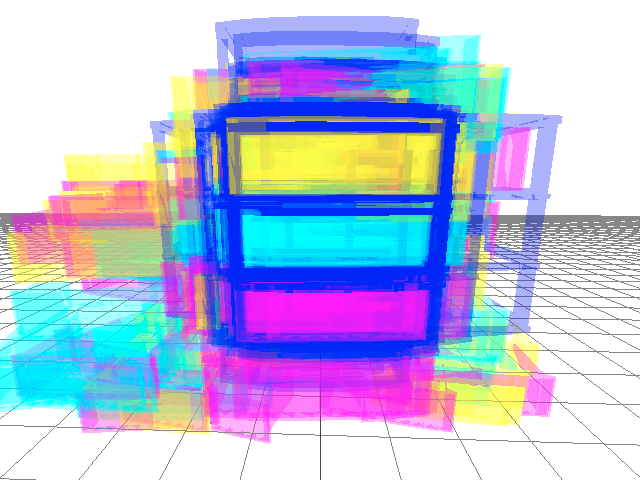}} ~
    \subfloat{\includegraphics[width=0.19\textwidth]{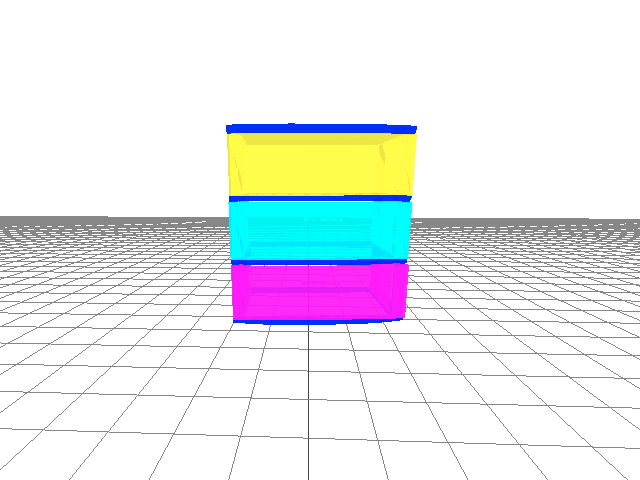}} ~
    \subfloat{\includegraphics[width=0.19\textwidth]{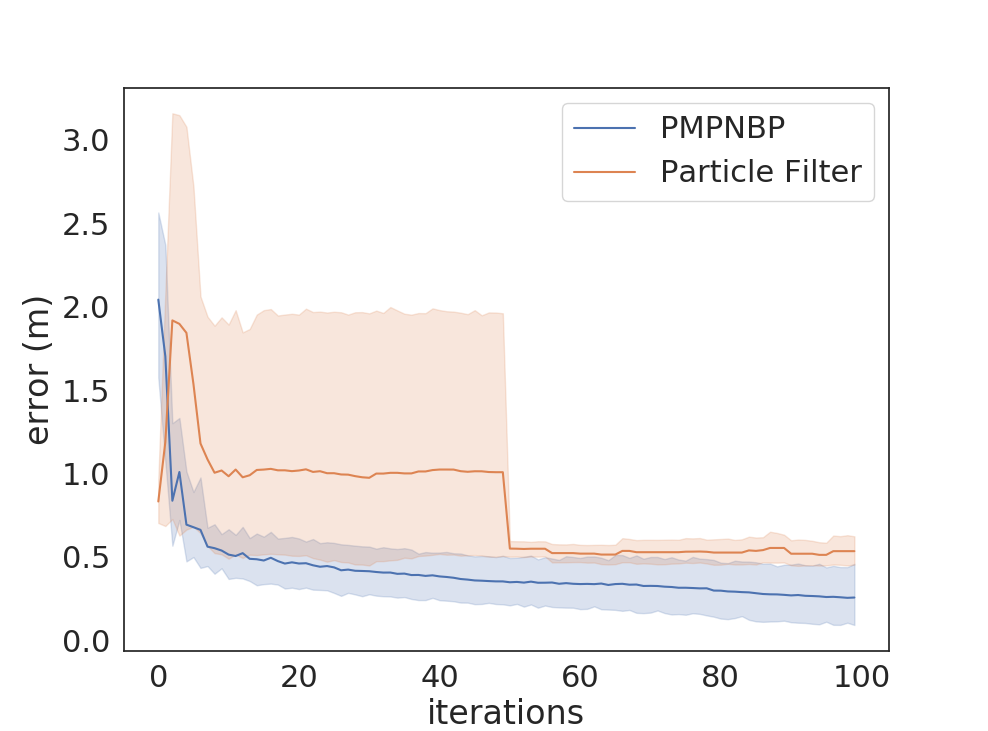}} \\
        
% 	\subfloat{\includegraphics[width=0.19\textwidth]{./figures/rgb_c4.jpg}} ~
%     \subfloat{\includegraphics[width=0.19\textwidth]{./figures/ic_c4_init_bel.png}} ~
%     \subfloat{\includegraphics[width=0.19\textwidth]{./figures/ic_c4_final_bel.png}} ~
%     \subfloat{\includegraphics[width=0.19\textwidth]{./figures/ic_c4_final_est.png}} ~
%     \subfloat{\includegraphics[width=0.19\textwidth]{./figures/ic_c4_conv_plot.png}} \\
        
% 	\subfloat{\includegraphics[width=0.19\textwidth]{./figures/rgb_c4.jpg}} ~
%     \subfloat{\includegraphics[width=0.19\textwidth]{./figures/ic_c4_init_bel.png}} ~
%     \subfloat{\includegraphics[width=0.19\textwidth]{./figures/ic_c4_final_bel.png}} ~
%     \subfloat{\includegraphics[width=0.19\textwidth]{./figures/ic_c4_final_est.png}} ~
%     \subfloat{\includegraphics[width=0.19\textwidth]{./figures/ic_c4_conv_plot.png}} \\
 
 \caption{\footnotesize Convergence of pose estimation on two different scenes: the first column shows the RGB image of each scene, second to fourth columns show the convergence results of PMPNBP. The second column shows randomly initialized belief particles, the third column shows the belief particles after 100 iterations, and the fourth column shows the maximum likely estimates of each part. The fifth column shows the estimation error (0.95 confidence interval) using PMPNBP with respect to the baseline particle filter method across 10 runs (400 particles and 100 iterations each). It can be seen that the baseline suffers from local minimas while PMPNBP is able to recover from them effectively.}
 \label{conv_results}
\end{figure*}

\subsection{``Pull'' Message Update}
\label{pmpnbp}

Given the overview of Nonparametric Belief Propagation above in Section~\ref{nbp_overview}, %\chad{use latex ref and label to generate section numbers}, 
we now describe our ``pull'' message passing algorithm. We represent message as a set of pairs instead of triplets in Equation~\ref{eq:3} which is
\begin{equation}
m_{t\rightarrow s} = \{(w_{ts}^{(i)},\mu_{ts}^{(i)}): 1\leq i\leq M\}
\end{equation}
Similarly, the marginal belief is summarized as a sample set
\begin{equation}
bel_s^n(X_s) = \{\mu_{s}^{(i)}: 1\leq i\leq T\}
\end{equation}
where $T$ is the number of samples representing the marginal belief.
We assume that there is a marginal belief over $X_s$ as $bel_s^{n-1}(X_s)$ from the previous iteration. To compute the $m_{t\rightarrow s}^n(X_s)$, at iteration $n$, we initially sample $\{\mu_{ts}^{(i)}\}_{i=1}^M$ from the belief $bel_s^{n-1}(X_s)$. Pass these samples over to the neighboring nodes $\rho(t)\setminus s$ and compute the weights $\{w_{ts}^{(i)}\}_{i=1}^M$. This step is described in \ref{alg:message_update}. The computation of $bel_s^{n}(X_s)$ is described in ~\ref{alg:belief_update}. The key difference between the ``push'' approach of the earlier methods (NBP~\cite{sudderth2003nonparametric} and PAMPAS~\cite{isard2003pampas}) and our ``pull'' approach is the message $m_{t\rightarrow s}$ generation. In the ``push'' approach, the incoming messages to $t$ determines the outgoing message $t\rightarrow s$. Whereas, in the ``pull'' approach, samples representing $s$ are drawn from its belief $bel_s$ from previous iteration and weighted by the incoming messages to $t$. This weighting strategy is computationally efficient. Additionally, the product of incoming messages to compute $bel_s$ is approximated by a resampling step as described in ~\ref{alg:belief_update}.

\subsection{Potential functions}
\subsubsection{Unary potential} \label{unary_potential}
Unary potential $\phi_t(X_t, Y_t)$ is used to model the likelihood by measuring how a pose $X_t$ explains the point cloud observation  $P_t$. The hypothesized object pose $X_t$ is used to position the given geometric object model and generate a synthetic point cloud $P_t^*$ that can be matched with the observation $P_t$. The synthetic point cloud is constructed using the object-part's geometric model available {\it a priori}. The likelihood is calculated as
\begin{equation}\label{eq:5}
    \phi_{t}(X_t, Y_t) = e^{\lambda_{r}d(P_t, P_t^*)}
\end{equation}
where $\lambda_{r}$ is the scaling factor, $d(P_t, P_t^*)$ is the sum of 3D Euclidean distance between the observed point $p \in P_t$ and rendered point $p^* \in P_t^*$ at each pixel location in the region of interest.

\subsubsection{Pairwise potential and sampling} \label{pairwise}
Pairwise potential $\psi_{t,s}(X_t|X_s)$ gives information about how compatible two object poses are given their joint articulation constraints captured by the edge between them. As mentioned in the Section\ref{problem_statement}, these constraints are captured using dual quaternions. Most often, the joint articulation constraints have minimum and maximum range in either prismatic or revolute types. We capture this information from URDF to get $R_{t|s} = [dq_{t|s}^a$, $dq_{t|s}^b]$ giving the limits of articulations. For a given $X_s$ and $R_{t|s}$, we find the distance between $X_t$ and the limits as $A=d(X_t, dq_{t|s}^a)$ and $B=d(X_t, dq_{t|s}^b)$, as well as the distance between the limits $C=d(dq_{t|s}^a, dq_{t|s}^b)$. Using a joint limit kernel parameterized by $(\sigma_{pos}, \sigma_{ori})$, we evaluate the pairwise potential as:
\begin{equation}\label{eq:6}
    \psi_{t,s}(X_t|X_s) = e^{-\frac{(A_{pos}+B_{pos}-C_{pos})^2}{2(\sigma_{pos})^2} -\frac{(A_{ori}+B_{ori}-C_{ori})^2}{2(\sigma_{ori})^2} }
\end{equation}
The pairwise sampling uses the same limits $R_{t|s}$ to sample for $X_t$ given a $X_s$. We uniformly sample a dual quaternion $\bar{X_t}$ that is between $[dq_{t|s}^a, dq_{t|s}^b]$ and transform it back to the $X_s$'s current frame of reference by $X_t=X_s*\bar{X_t}$. 

\section{Experiments and Results}
\begin{figure*}[t!] \label{partial_obs}
    \centering
    \captionsetup[subfigure]{labelformat=empty}
    
%   \subfloat{\includegraphics[width=0.2\textwidth]{./figures/rgb_spl_3.jpg}} ~
%     \subfloat{\includegraphics[width=0.2\textwidth]{./figures/ic_teaser_pc}} ~
%     \subfloat{\includegraphics[width=0.2\textwidth]{./figures/ic_teaser_final_est}} ~    
%     \subfloat{\includegraphics[width=0.2\textwidth]{./figures/ic_teaser_final_est}} \\
	\subfloat{\includegraphics[width=0.245\textwidth]{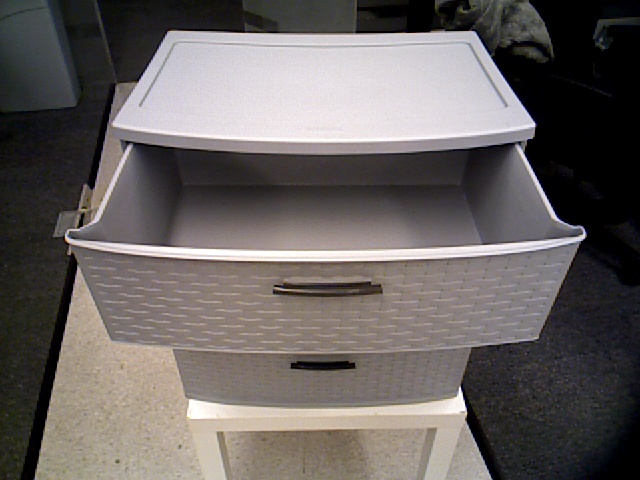}} ~
    \subfloat{\includegraphics[width=0.245\textwidth]{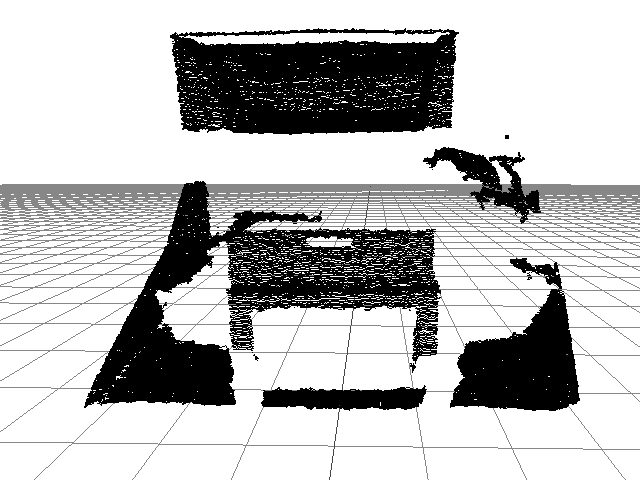}} ~
    \subfloat{\includegraphics[width=0.245\textwidth]{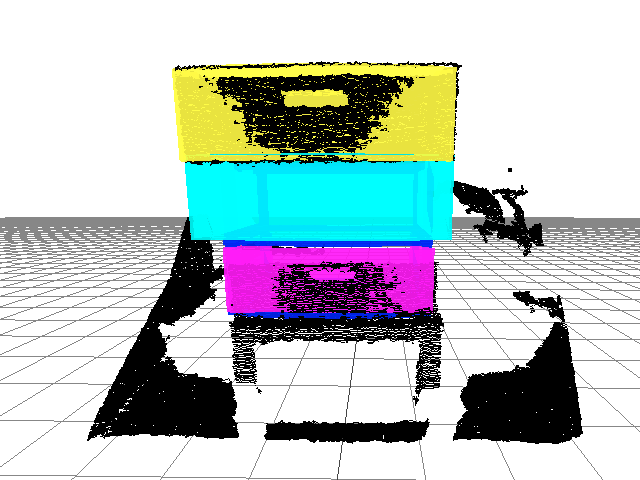}} ~    
    \subfloat{\includegraphics[width=0.245\textwidth]{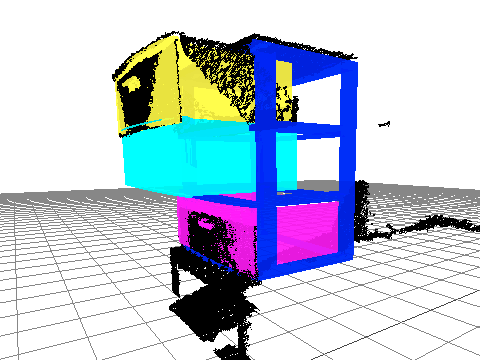}} \\
	\subfloat{\includegraphics[width=0.245\textwidth]{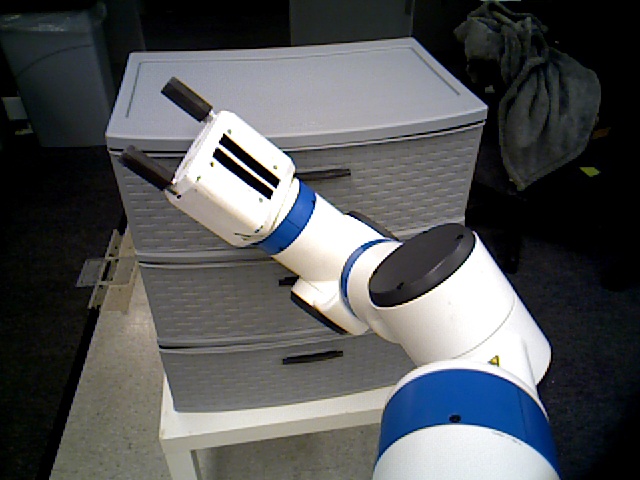}} ~
    \subfloat{\includegraphics[width=0.245\textwidth]{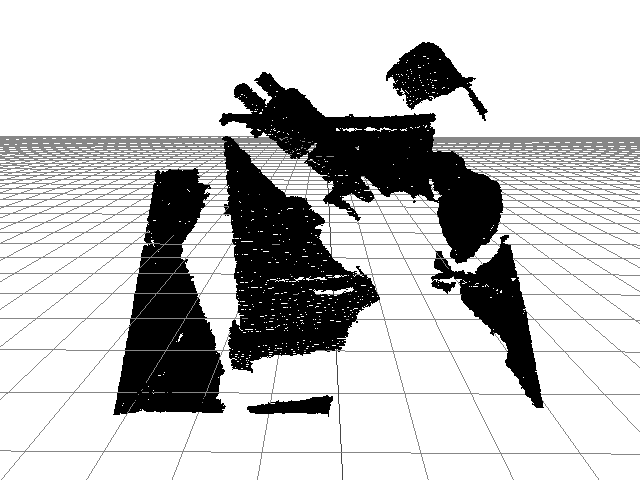}} ~
    \subfloat{\includegraphics[width=0.245\textwidth]{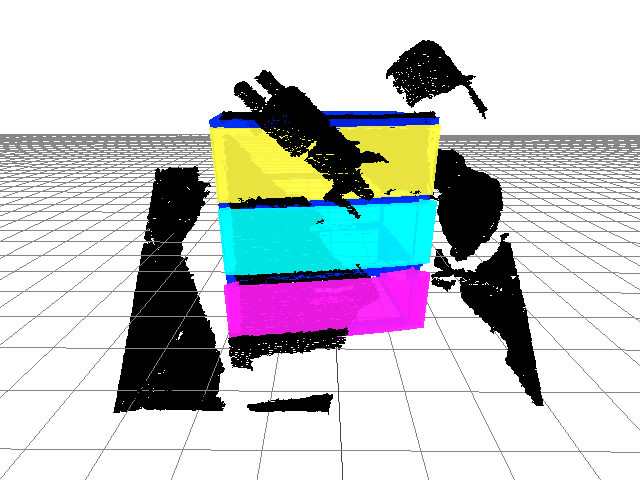}} ~    
    \subfloat{\includegraphics[width=0.245\textwidth]{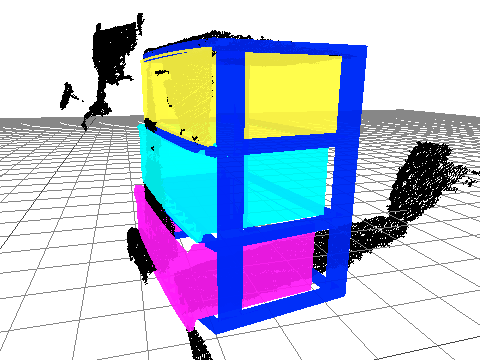}} \\    
  \subfloat[Original Scene]{\includegraphics[width=0.245\textwidth]{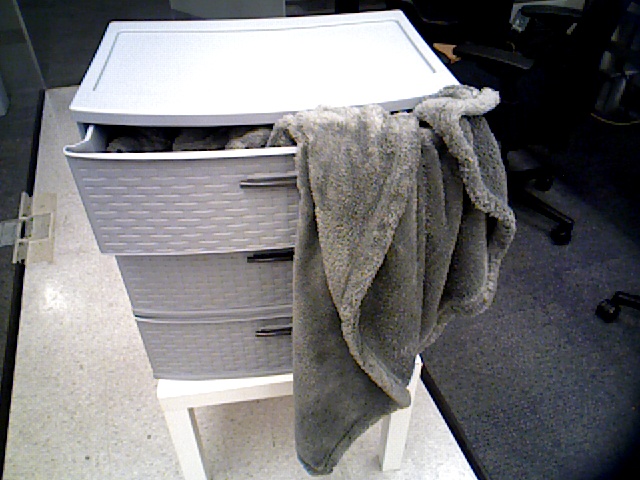}} ~
    \subfloat[Incomplete Observation]{\includegraphics[width=0.245\textwidth]{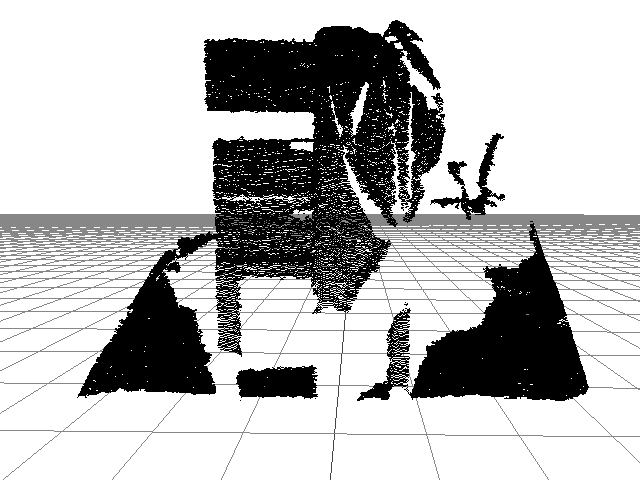}} ~
    \subfloat[MLE using PMPNBP]{\includegraphics[width=0.245\textwidth]{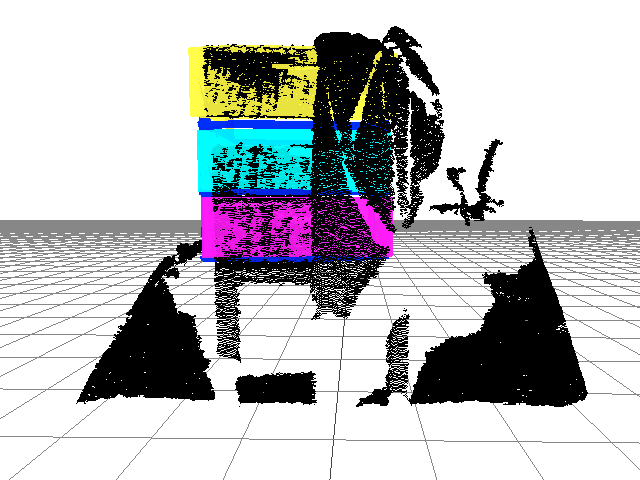}} ~    
    \subfloat[MLE from a different view]{\includegraphics[width=0.245\textwidth]{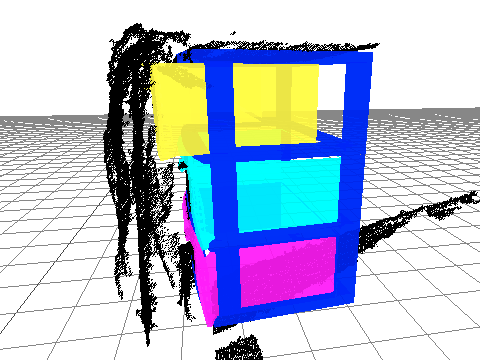}} 
%     \\
%   \subfloat[Original Scene]{\includegraphics[width=0.24\textwidth]{./figures/rgb_spl_4.jpg}} ~
%     \subfloat[Incomplete Observation]{\includegraphics[width=0.24\textwidth]{./figures/ic_spl_4_pc.png}} ~
%     \subfloat[MLE using PMPNBP]{\includegraphics[width=0.24\textwidth]{./figures/ic_spl_4_est.png}} ~    
%     \subfloat[MLE from a different view]{\includegraphics[width=0.24\textwidth]{./figures/ic_spl_4_v_est.png}}
 \caption{\footnotesize Partial and incomplete observations due to self and environmental occlusions are handled by PMPNBP in estimating plausible pose with accuracy}
 \label{incomp_obs}
\end{figure*}
\subsection{Experimental setup}
We use Fetch robot, a mobile manipulation platform for our data collection and manipulation experiments. RGBD data is collected using an ASUS Xtion RGBD sensor mounted on the robot along with the intrinsic and camera to robot base transform. We use CUDA-OpenGL interoperation to render synthetic scenes on large set of poses in a single render buffer on a GPU. We render scenes as depth images, then project them back to 3D point clouds via camera intrinsic parameters. %For manipulation, we use TRAC-IK~\cite{beeson2015trac} for inverse kinematics and MoveIt!~\cite{sucanmoveit} to perform motion planning afterwards. 

\subsection{Articulated Objects Models}
We used a cabinet with three drawers as our articulated object in the experiment. CAD model of the object is obtained from the Internet and annotation of their articulations are performed on Blender to generate URDF models. Obtaining geometrical models and articulation models can either be crowd-sourced ~\cite{gouravajhala2018eureca} or learned using human or robot interactions ~\cite{martin14online}.

\subsection{Baseline}
We implemented Monte Carlo localization (particle filter) method that has object specific state representation. For example, the Cabinet with 3 drawers have state representation of $(x, y, z, \phi, \psi, \chi, t_a, t_b, t_c)$ where the first 6 elements describe the 6D pose of the object in the world and ${t_a, t_b, t_c}$ represent the prismatic articulation. The measurement model in the implementation uses the unary potential described in the Section \ref{unary_potential}. Instead of rendering a point cloud of each object-part, the entire object in the hypothesized pose is rendered for measuring the likelihood. As the observations are static, the action model in the standard particle filter is replaced with a Gaussian diffusion over the object poses.

\begin{figure*}[t!] \label{partial_obs}
    \centering
    \captionsetup[subfigure]{labelformat=empty}

% 	\subfloat[Fetch robot]{\includegraphics[width=0.25\textwidth]{./figures/robot.png}} ~~
    \subfloat[(a) Fetch robot in original scene]{\includegraphics[width=0.28\textwidth]{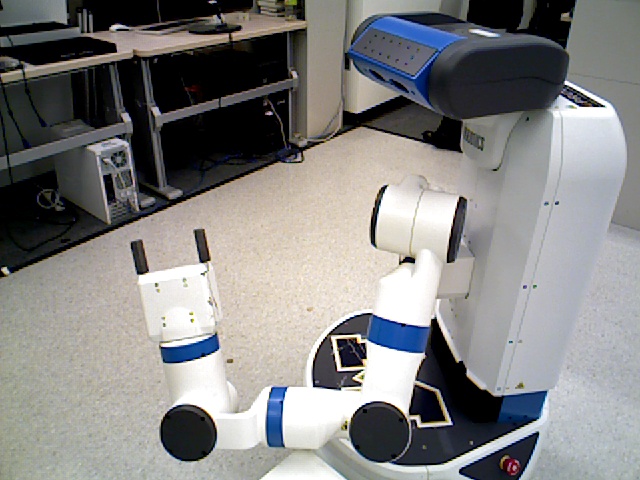}} ~~
    \subfloat[(b) Point cloud observation]{\includegraphics[width=0.28\textwidth]{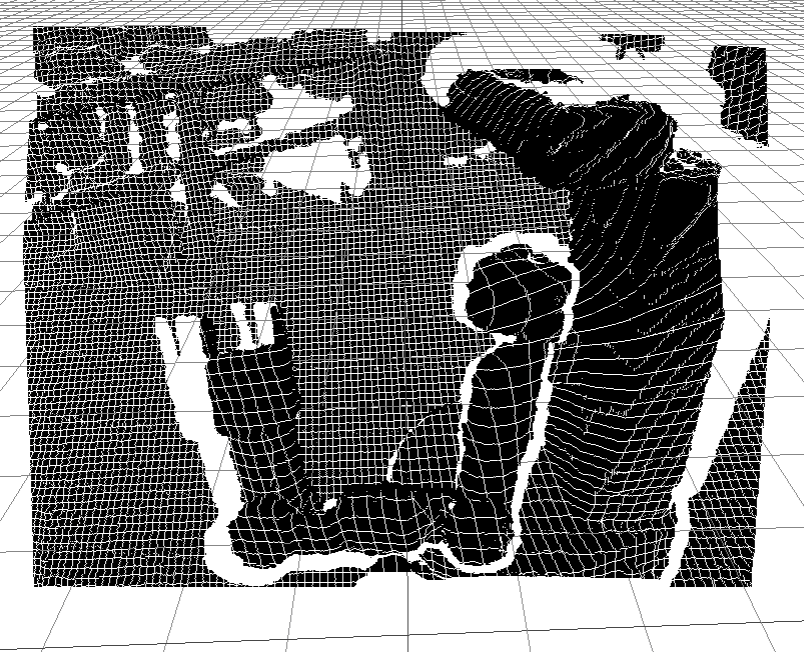}} ~~
    \subfloat[(c) MRF model]{\includegraphics[width=0.28\textwidth]{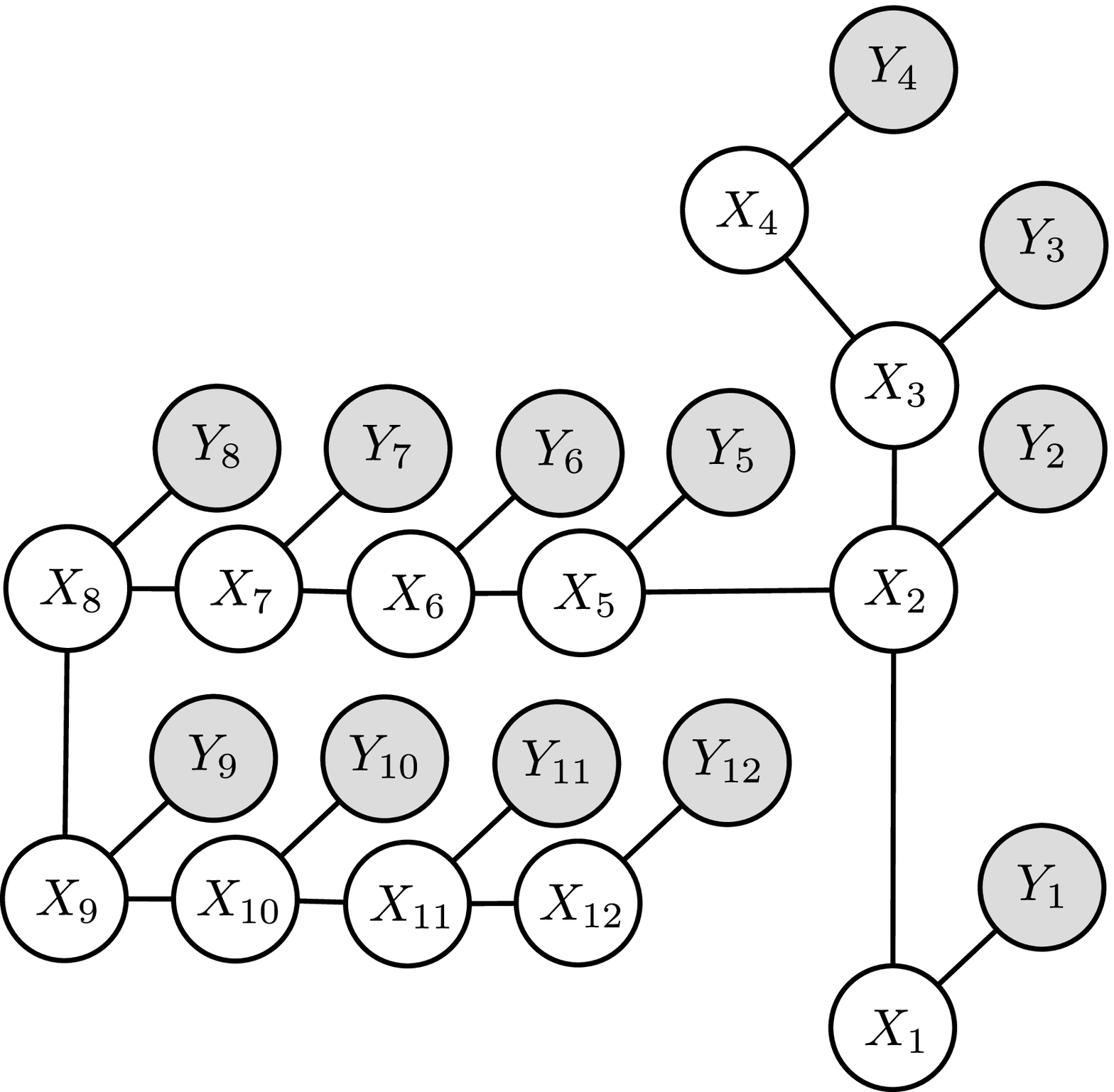}} \\
    
    \subfloat[(d) Belief at iteration 1  (view 1)]{\includegraphics[width=0.28\textwidth]{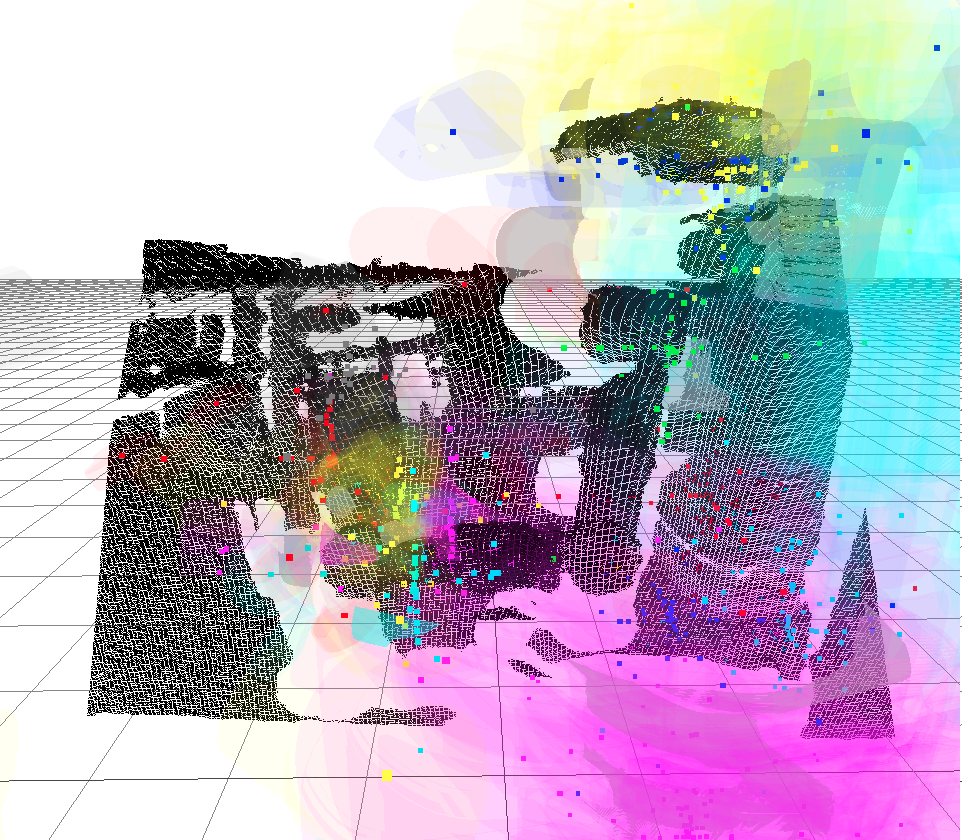}} ~~
    \subfloat[(e) Belief at iteration 1000 (view 1)]{\includegraphics[width=0.28\textwidth]{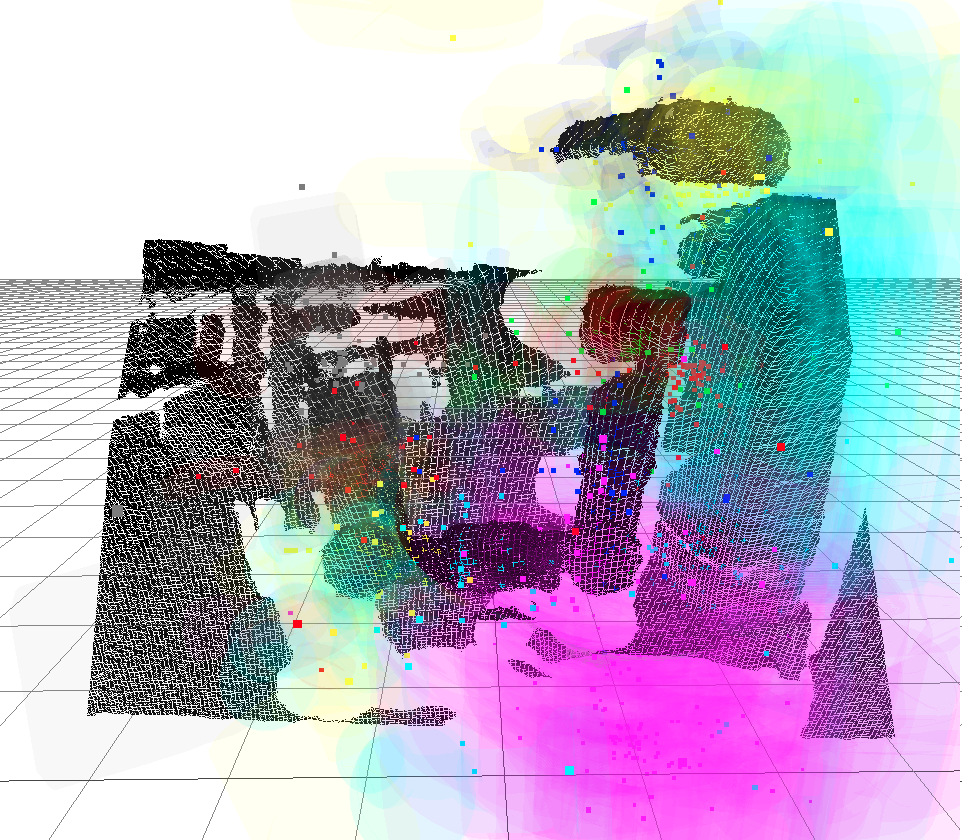}} ~~
    \subfloat[(f) MLE (view 1)]{\includegraphics[width=0.28\textwidth]{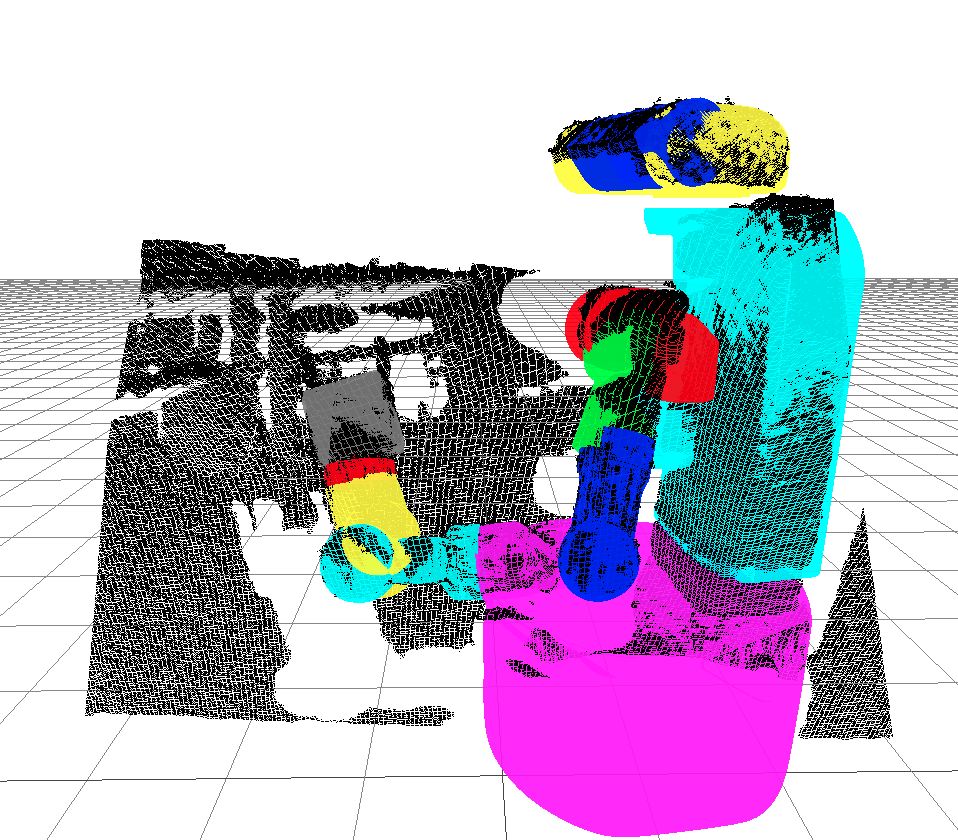}} \\
    
    \subfloat[(g) Belief at iteration 1 (view 2)]{\includegraphics[width=0.28\textwidth]{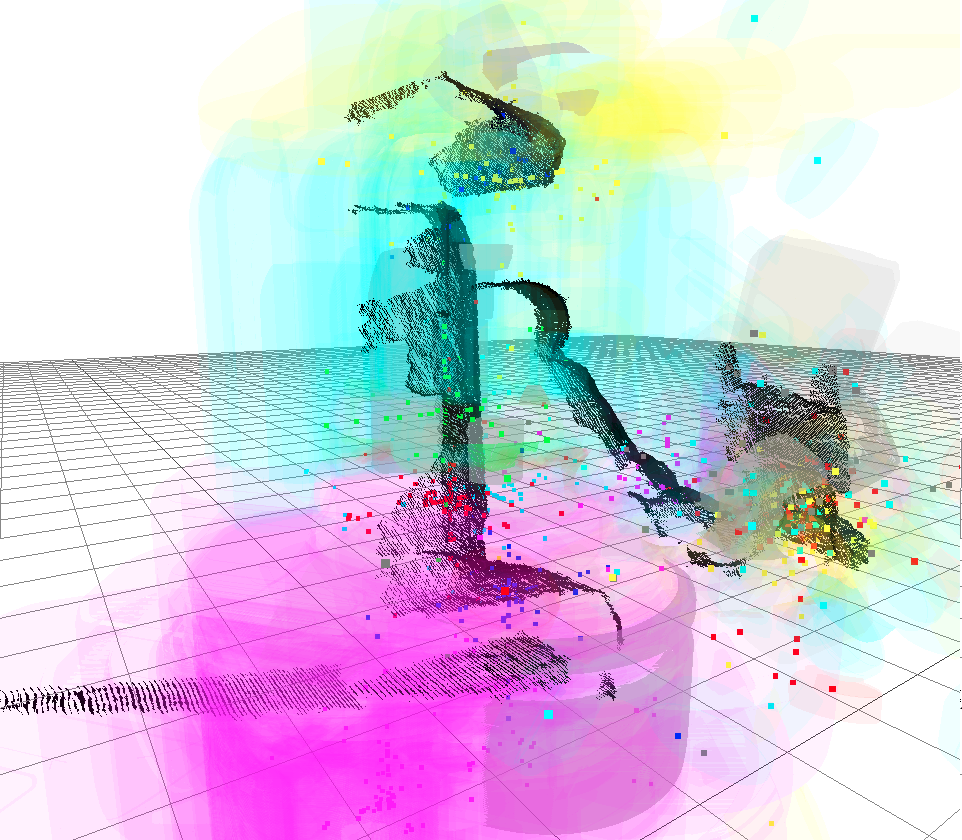}} ~~
    \subfloat[(h) Belief at iteration 1000 (view 2)]{\includegraphics[width=0.28\textwidth]{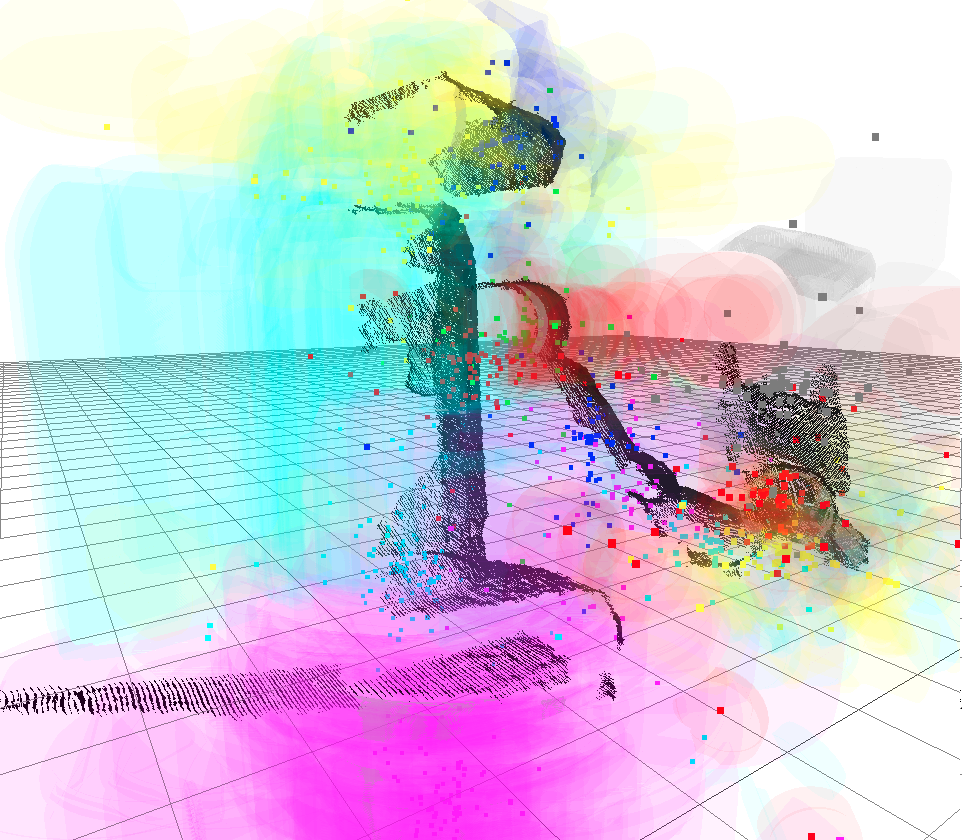}} ~~
    \subfloat[(i) MLE (view 2)]{\includegraphics[width=0.28\textwidth]{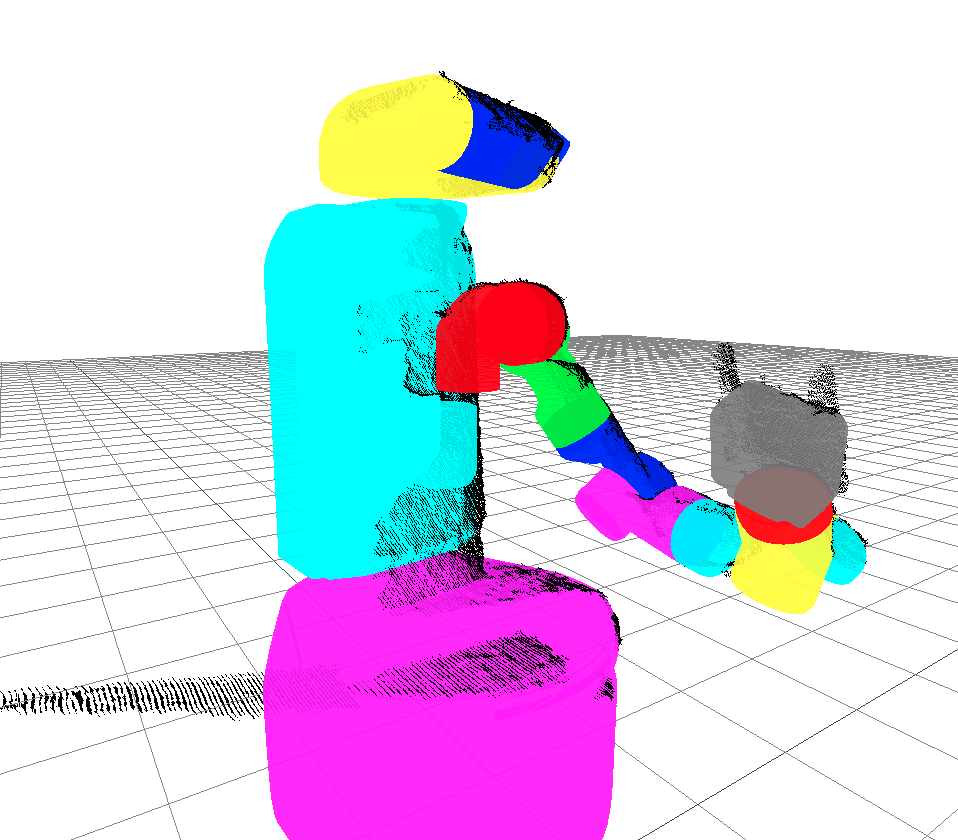}} \\
 \caption{\footnotesize Factored pose estimation using PMPNBP extends to articulated objects such as Fetch robot (a) which has 12 nodes and 11 edges in the probabilistic graphical model (c). For a scene (a), which has partial 3D point cloud observation (b), the PMPNBP message passing algorithm, propagates the belief samples from iteration 1 (d and g) to iteration 1000 (e and h), that leads to MLE (f and i).}
 \label{robot_obs}
\end{figure*}

\begin{figure*}[!t]
    \centering
    \captionsetup[subfigure]{labelformat=empty}
% 	\subfloat[]{\includegraphics[width=0.8\textwidth]{./figures/ic_last1.PNG}} 
	\subfloat[(a) Scene observed]{\includegraphics[width=0.28\textwidth]{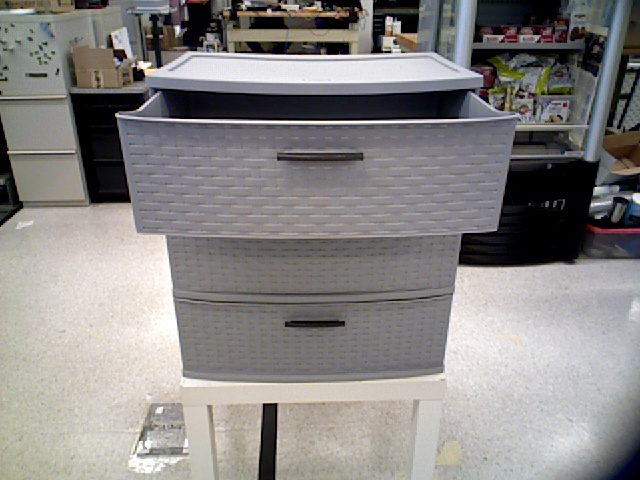}} ~    
    \subfloat[(b) Estimate from PMPNBP]{\includegraphics[width=0.28\textwidth]{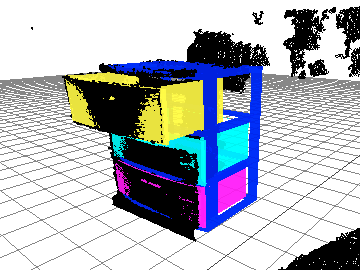}} ~
    \subfloat[(c) Confidence ellipsoids on belief samples]{\includegraphics[width=0.28\textwidth]{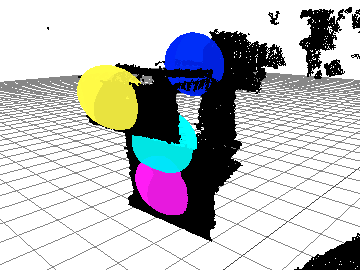}} \\
    \subfloat[(d) Grasping drawer 3]{\includegraphics[width=0.3\textwidth]{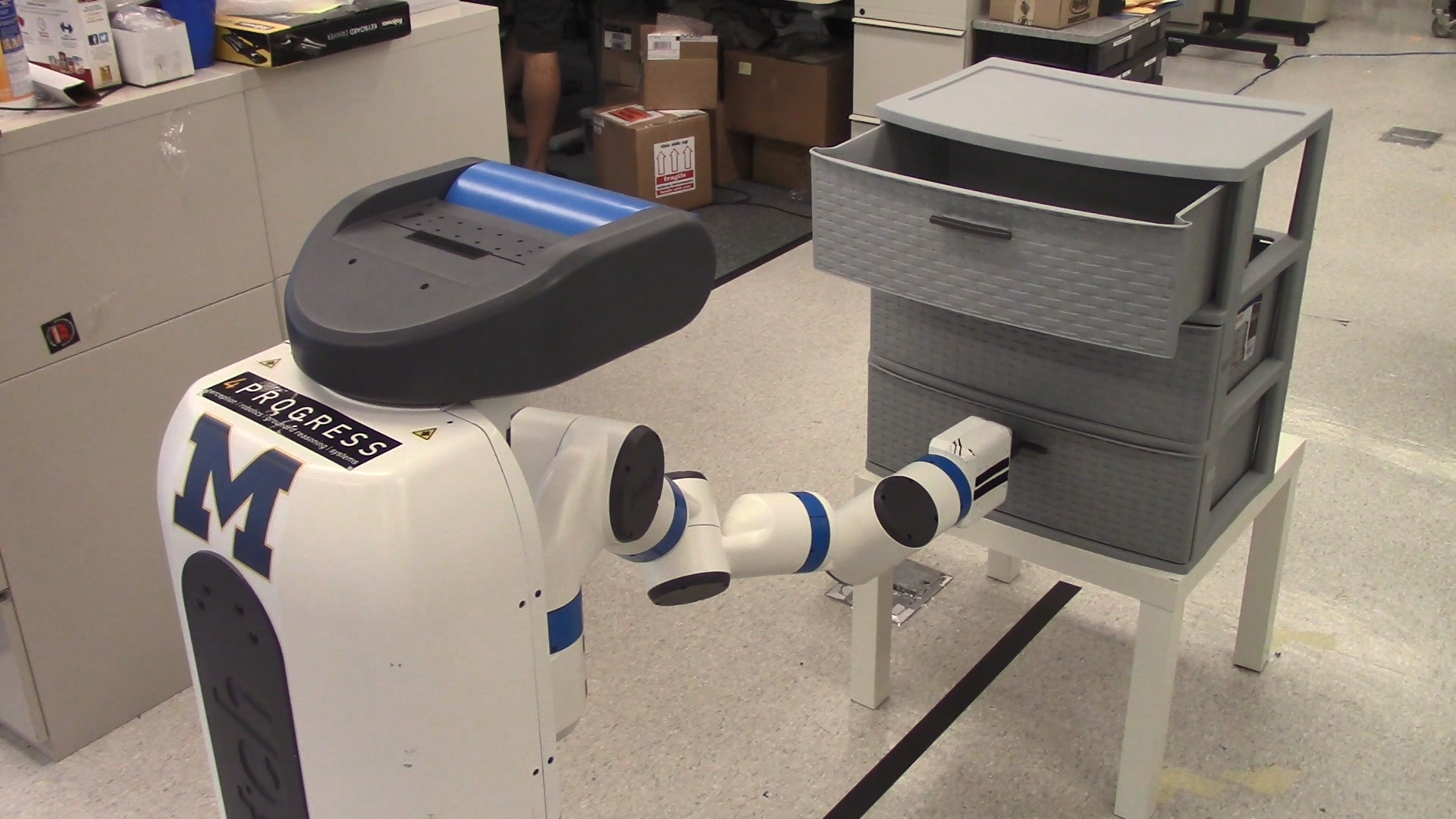}} ~
    \subfloat[(e) Opening drawer 3]{\includegraphics[width=0.3\textwidth]{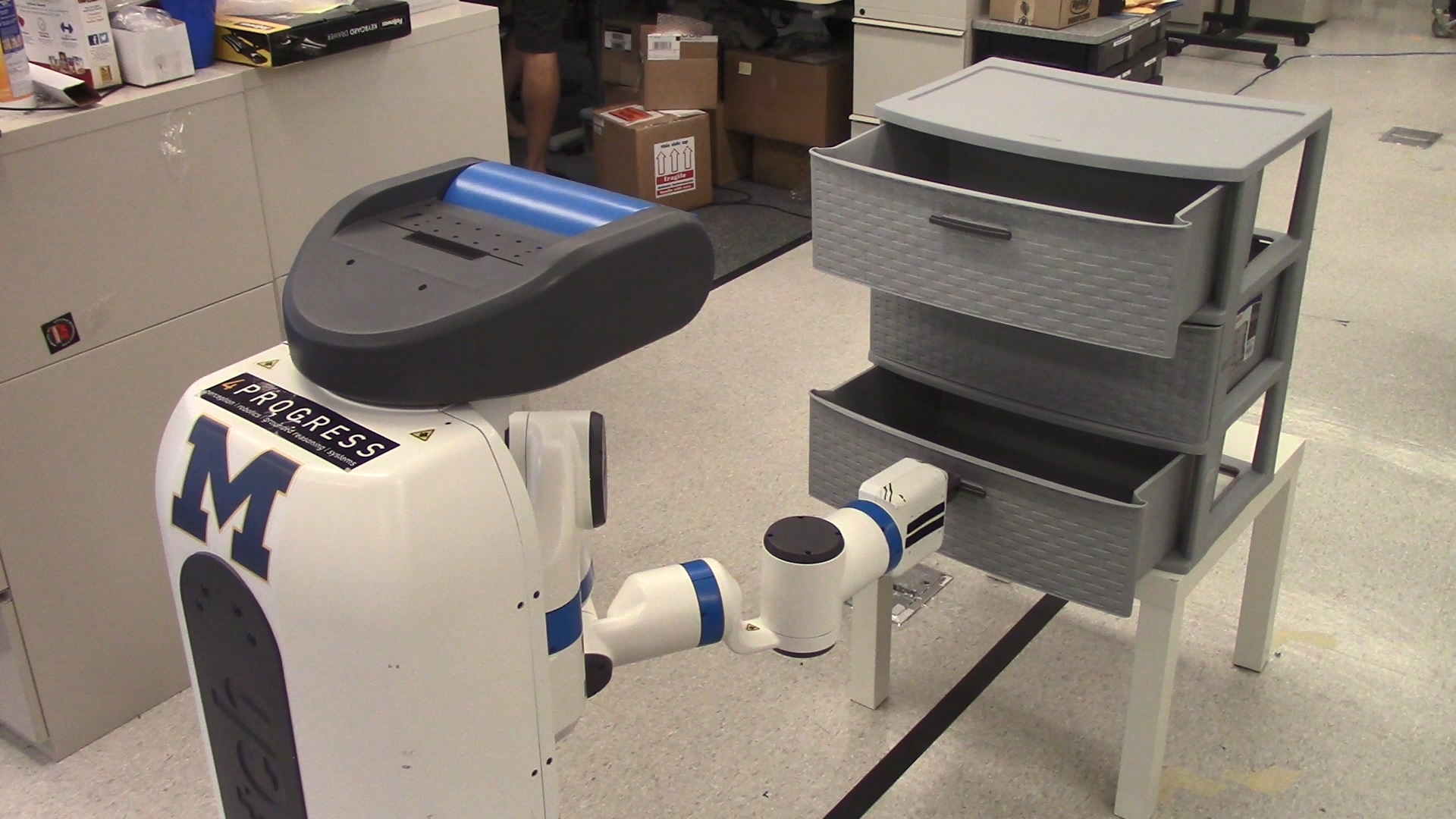}} \\

%   \subfloat[World scene]{\includegraphics[width=0.19\textwidth]{./figures/second_idle.jpg}} ~
% 	\subfloat[Scene observed]{\includegraphics[width=0.19\textwidth]{./figures/rgb_re_2.jpg}} ~    
%     \subfloat[Estimate from PMPNBP]{\includegraphics[width=0.19\textwidth]{./figures/ic_re_2_est.png}} ~
% %     \subfloat[Variance in estimation(TODO)]{\includegraphics[width=0.2\textwidth]{./figures/ic_teaser_final_est}} \\
%     \subfloat[Scene observed after moving the camera]{\includegraphics[width=0.19\textwidth]{./figures/rgb_re_3.jpg}} ~    
%     \subfloat[Estimate from PMPNBP]{\includegraphics[width=0.19\textwidth]{./figures/ic_re_3_est.png}} \\
% %     \subfloat[Variance in estimation(TODO)]{\includegraphics[width=0.2\textwidth]{./figures/ic_teaser_final_est}} \\
%     \subfloat[Grasping drawer 3]{\includegraphics[width=0.19\textwidth]{./figures/second_grasp.jpg}} ~
%     \subfloat[Opening drawer 3]{\includegraphics[width=0.19\textwidth]{./figures/second_open.jpg}} \\
    
 \caption{\footnotesize Robot Experiments 1: The task for the robot is to open the drawer 3 (bottom) while having the drawer 1 open. The robot estimates the state of the object with certainty as shown in (b) with drawer 1 open and drawer 3 closed. In addition to the estimate, covariance can be calculated (shown as ellipsoids in (c) with 75\% confidence interval). This could be used to decide that the estimation is certain with a threshold on the standard deviation on each of the dimensions of the pose. In this case the standard deviation of ($x, y, z$) falls below the threshold 0.25cm, and allows the robot to perform the opening action. (d-e) shows the robot performing opening action on drawer 3 using the estimate.} %\kar{Show this very clearly - what is belief, how is the covariance calculated (along with visualization), and enlarge the picture for easy understanding}}.
 \label{robot_exp1}
\end{figure*}

\subsection{Convergence Results}
In the Figure.~\ref{conv_results}, we show the convergence of the proposed method visually for two scenes containing different point cloud observations. We collected point cloud observations of the objects in arbitrary poses and performed inference using both the proposed PMPNBP and the baseline Monte Carlo localization. Entire point cloud observed by the sensor is used as the observation for all the object-parts. The first column shows the scene (RGB not used in the inference). Second column shows the uniformly initialized poses of the object-parts on the entire point cloud. Third column shows the propagated belief particles for each object-part after 100 iterations. Fourth column shows the Maximum Likely Estimate (MLE) of each object-part using the belief particles from the third column. 

%This plot indicates the efficiency (TODO). 
For the results shown in Figure.~\ref{conv_results}, we ran our inference for 100 iterations with 400 particles representing the messages. 10 different runs are used to generate the convergence plot that shows the mean  and variance in error across the runs. We adopt the average distance metric (ADD) proposed in \cite{hinterstoisser2012model, xiang2017posecnn} for the evaluation. The point cloud model of the object-part is transformed to its ground truth dual quaternion ($dq$) and to the estimated pose's dual quaternion ($\bar{dq}$). Error is calculated as the pointwise distance of these transformation pairs normalized by the number of points in the model point cloud.
\begin{equation}\label{r_eq:1}
ADD = \frac{1}{m} \sum_{p\in \mathcal{M}} \| \bar{dq}*p*\bar{dq_c} - dq*p*dq_c \|
\end{equation}
where ($\bar{dq_c}$) and ($dq_c$) are the conjugates of the dual quaternions ~\cite{gilitschenski2014new,kenwright2012beginners}, $m$ is the number of 3D points in the model set $\mathcal{M}$. 

% \begin{figure}
%     \centering
%     \captionsetup[subfigure]{labelformat=empty}
% 	\subfloat[No bounding box]{\includegraphics[width=0.49\columnwidth]{./figures/rgb_c4.jpg}}~
%     \subfloat[With bounding box]{\includegraphics[width=0.49\columnwidth]{./figures/rgb_c4.jpg}}\\
%     \subfloat[dummy plot]{\includegraphics[width=1.0\columnwidth]{./figures/prettified_grouped_bar.png}}
% \caption{Influence of prior models:}
%  \label{prior_models}
% \end{figure}

% \subsection{Influence of prior models}
% The proposed approach can benefit from prior appearance learned models to detect the object in the scene and provide a region of interest for initializing the belief samples. This will increase the convergence rate and narrow down the search space. However if the learned models provided noisy priors, the probabilistic nature of PMPNBP is capable of overcoming these limitations. In Figure.~\ref{prior_models}, we show the three different observations (i.e. no region of interest, region of interest and noisy region of interest) and evaluate the error in the estimates on different scenes that contain the articulated objects. 

\subsection{Partial and incomplete observations}
Articulated models suffer from self-occlusions and often environmental occlusions. By exploiting the articulation constraints of an object in the pose estimation, our inference method is able to estimate a physically plausible estimate that can explain the partial or incomplete observations. In Figure.~\ref{incomp_obs} we show three compelling cases that indicates the strength of our inference method. In the first case, the drawer 1 heavily occludes the bottom drawers resulting in limited observations on drawer 2 and 3. PMPNBP is able to estimate a plausible pose given the constraints. In the second case, the cabinet is occluded by the robot's arm, while in the third case, a blanket from the drawer 1 occludes half of the object. PMPNBP is able to recover from these occlusions and produce a plausible estimate along with belief of possible poses. 

The factored approach proposed in this paper scales to objects with higher number of links and joints with combinations of articulations. This is evaluated by estimating the pose of a Fetch robot that has 12 nodes and 11 edges in its graphical model. The graphical model is constructed using the URDF model of the robot. This is shown in Figure.~\ref{robot_obs}(c) where the robot is observed using the a depth camera. Figure.~\ref{robot_obs}(a \& b) show the original scene and its point cloud observation with partial sensor data on the base, torso and the head of the robot. PMPNBP is able to estimate the pose of the robot by iteratively passing messages for 1000 iterations. Figures.~\ref{robot_obs}(d-f) and Figures.~\ref{robot_obs}(g-i) show the belief samples of the robot links at iteration 1 and 1000 followed by the most likely estimation (MLE) from two different view points for better visualization.

\subsection{Benefits of maintaining belief towards planning actions}
\begin{figure*}[t!]
    \centering
    \captionsetup[subfigure]{labelformat=empty}
    % \subfloat[]{\includegraphics[width=0.87\textwidth]{./figures/ic_last2.PNG}}
% 	\subfloat[Scene observed]{\includegraphics[width=0.19\textwidth]{./figures/rgb_re_1.jpg}} ~    
%     \subfloat[Estimate from PMPNBP]{\includegraphics[width=0.19\textwidth]{./figures/ic_re_1_est.png}} ~
% %     \subfloat[Variance in estimation(TODO)]{\includegraphics[width=0.2\textwidth]{./figures/ic_teaser_final_est}} \\
%     \subfloat[Grasping drawer 3]{\includegraphics[width=0.19\textwidth]{./figures/first_grasp.jpg}} ~
%     \subfloat[Opening drawer 3]{\includegraphics[width=0.19\textwidth]{./figures/first_open.jpg}} \\

%   \subfloat[World scene]{\includegraphics[width=0.19\textwidth]{./figures/second_idle.jpg}} ~
	\subfloat[(a) Scene observed]{\includegraphics[width=0.28\textwidth]{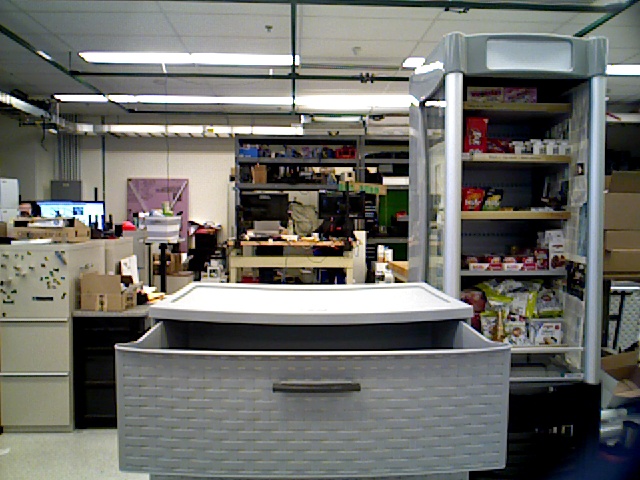}} ~    
    \subfloat[(b) Estimate from PMPNBP]{\includegraphics[width=0.28\textwidth]{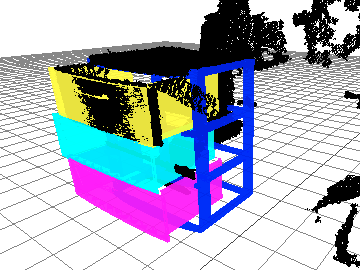}} ~
     \subfloat[(c) Confidence ellipsoids on belief samples]{\includegraphics[width=0.28\textwidth]{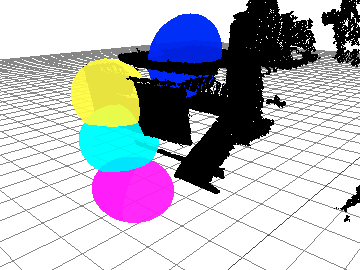}} \\
    \subfloat[(d) Scene observed after moving the camera]{\includegraphics[width=0.28\textwidth]{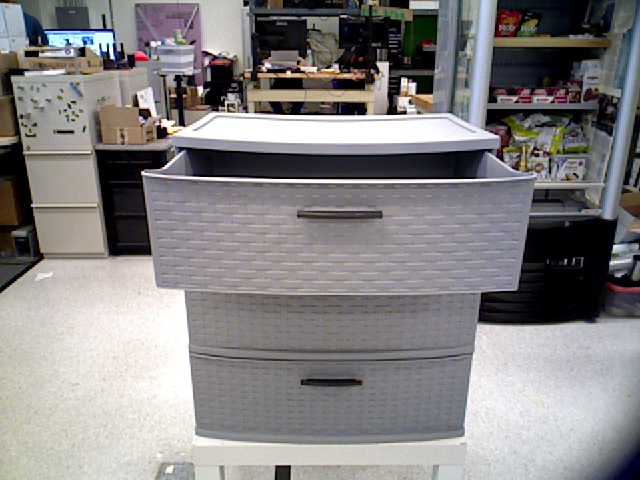}} ~    
    \subfloat[(e) Estimate from PMPNBP]{\includegraphics[width=0.28\textwidth]{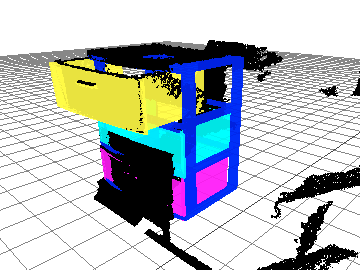}} ~
     \subfloat[(f) Confidence ellipsoids on belief samples]{\includegraphics[width=0.28\textwidth]{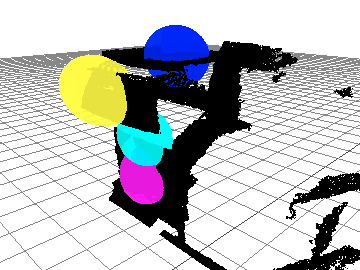}} \\
    \subfloat[(g) Grasping drawer 3]{\includegraphics[width=0.28\textwidth]{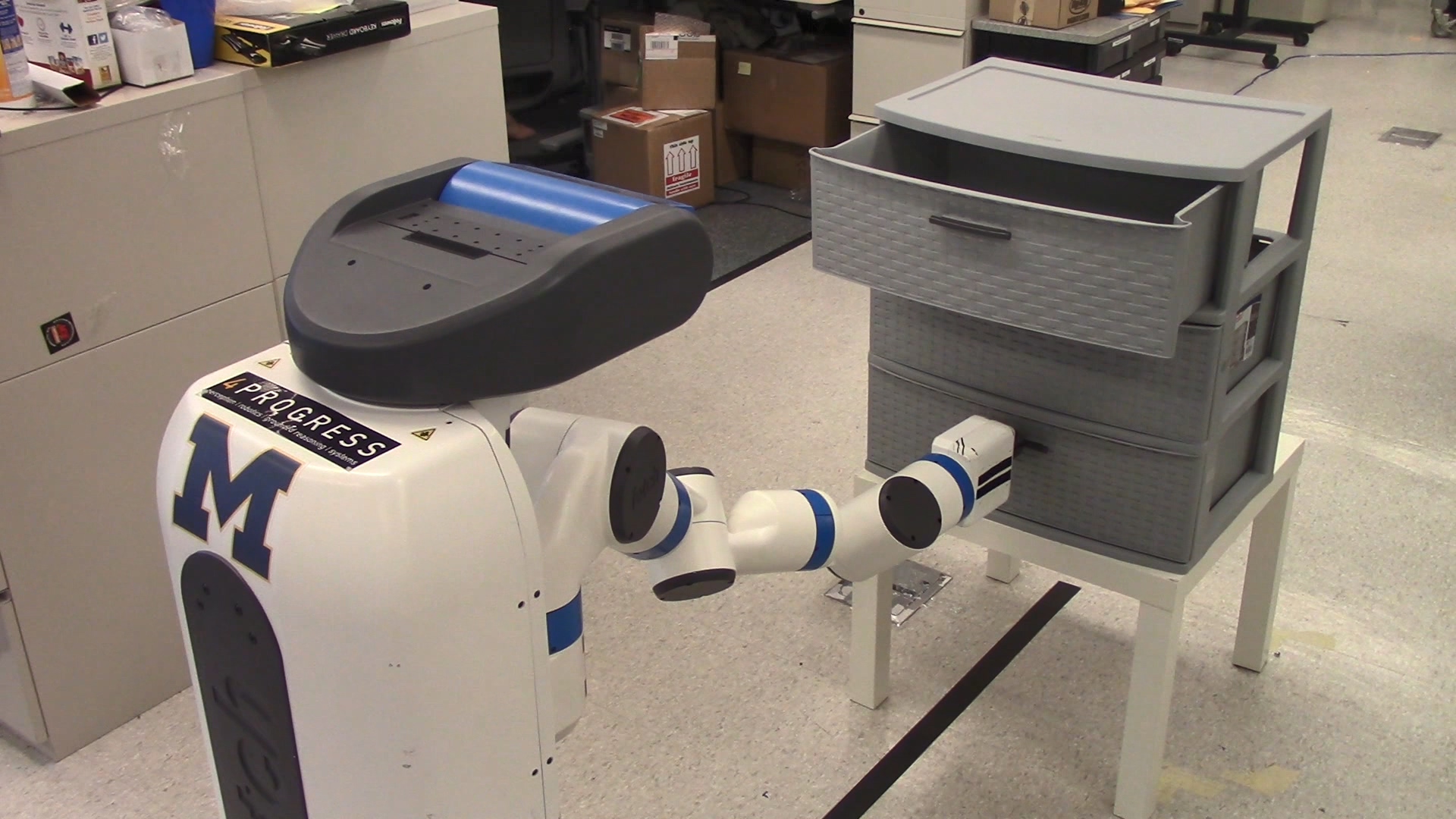}} ~
    \subfloat[(h) Opening drawer 3]{\includegraphics[width=0.28\textwidth]{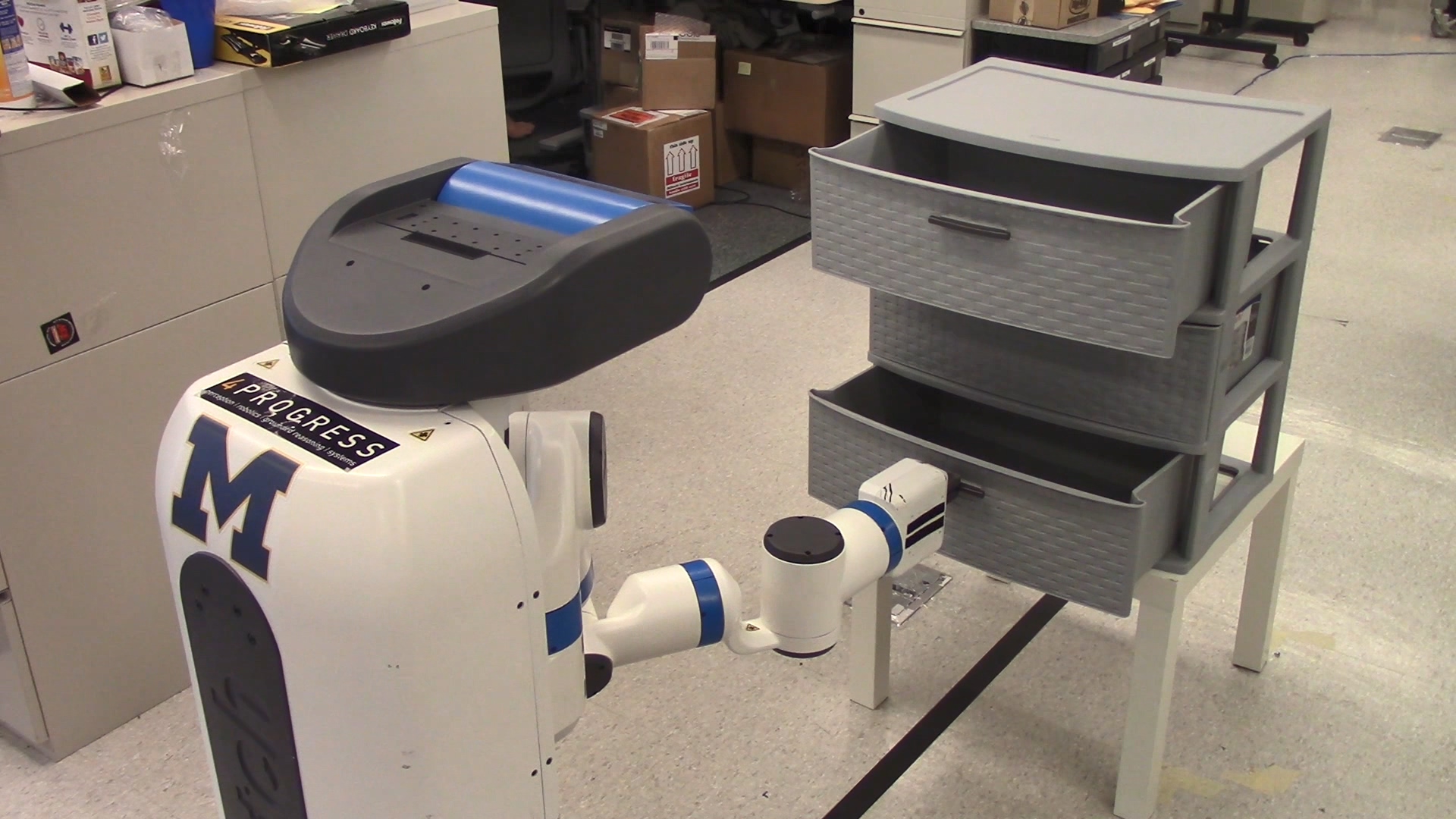}} \\
    
 \caption{\footnotesize Robot Experiments 2: The task remains the same as Figure.~\ref{robot_exp1}. However the robot is not directly observing the drawer 3 (a). The inference using PMPNBP gives an estimate (b). However, the standard deviation in the position (as seen in (c) with ellipsoids of larger radii), is higher than the threshold 0.25cm. To reduce the uncertainty in the estimation, the robot takes an intermediate action (changing the viewpoint) to gain more information about the scene. This results in new scene (d). Running inference using PMPNBP on this scene gives an estimate (e) with covariance ellipsoids as shown in (f). This satisfies the threshold on the standard deviations, enabling the robot to perfom the (g-h) grasping and opening of drawer 3.} 
\label{robot_exp2}
\end{figure*}

We show how the belief propagation approach aids in planning with a simple task illustration. Assume that the robot is performing a larger task of storing elements into the drawer 3. In a subtask, the goal is to open the drawer 3. With this setting (see Figure.~\ref{robot_exp1}) the robot is perceiving the current scene by estimating the pose of the cabinet, along with covariance on the belief for each part. We set a maximum threshold of 0.25cm on the standard deviation of $(x, y, z)$ dimensions to decide if the estimation is certain or not. In this case, the standard deviation from the belief falls within this threshold and the robot is certain that the drawer 1 is open and drawer 3 is closed. Hence, the robot performs opening drawer 3 action. For the same task but with a different observation (see Figure.~\ref{robot_exp2}), the robot estimates the pose of the cabinet, along with its covariance. However, in this case, the robot is not certain about the estimation as the standard deviation is bigger than the threshold. This enables the robot to take an intermediate action (of lowering its torso) that provides a new observation of the cabinet. With this new observation, the robot perceives that the drawer 3 is closed with more certainty and performs an open action. This is an illustration of how the belief can be used in planning actions. More rigorous experiments with the choice of thresholds for different objects and tasks will be detailed in the future work.

\section{Conclusion}
We proposed Pull Message Passing algorithm for Nonparametric Belief Propagation (PMPNBP), an efficient algorithm to estimate the poses of articulated objects. This problem was formulated as a graph inference problem for a Markov Random Field (MRF). We showed that the PMPNBP outperforms the baseline Monte Carlo localization method quantitatively. Qualitative results are provided to show the pose estimation accuracy of PMPNBP under a variety of occlusions. We also showed the scalability of the algorithm to articulated objects with higher number of nodes and edges in their probabilistic graphical models. In addition, we illustrated how belief propagation can benefit robot manipulation tasks. The notion of uncertainty in the inference is inevitable in robotic perception. Our proposed PMPNBP algorithm is able to accurately estimate the pose of articulated objects and maintain belief over possible poses that can benefit a robot in performing a task.

%%%%%%%%%%%%%%%%%%%%%%%%%%%%%%%%%%%%%%%%%%%%%%%%%%%%%%%%%%%%%%%%%%%%%%%%%%%%%%%%

\balance
\bibliographystyle{abbrv}

%\bibliography{nbp_rel}
\end{document}